\documentclass[journal]{IEEEtran}
\usepackage{amsmath}
\usepackage{cite}
\usepackage{multirow} 
\usepackage{multicol}
\usepackage{arydshln}
\usepackage{subfigure}
\usepackage{graphicx}
\usepackage{url}
\usepackage{amsmath}
\usepackage{amsfonts}
\usepackage{amssymb}
\usepackage[most]{tcolorbox}
\usepackage{ragged2e}
\usepackage{subfigure}
\usepackage{pgfplots}
\usepackage{booktabs}
\usepackage{longtable}
\usepackage{times}
\usepackage{arydshln}
\usepackage{booktabs}
\usepackage{float}
\usepackage{latexsym}
\usepackage{tabularx}
\usepackage[ruled, vlined, linesnumbered]{algorithm2e}
\usepackage{setspace}

\usepackage{pifont}% http://ctan.org/pkg/pifont
\usepackage{amsmath,amsfonts}
\usepackage{enumerate}
\usepackage{array}
\usepackage{wasysym}
\usepackage{url}
\usepackage{amssymb}
\usepackage{booktabs}
\usepackage{multirow}
\usepackage{threeparttable}
\usepackage[table]{xcolor} 
\usepackage{colortbl}
\usepackage{stfloats}
\usepackage{verbatim}
\usepackage{graphicx}
\usepackage{algpseudocode} 
\usepackage{enumitem}
\usepackage{makecell}
\usepackage{pifont}
\usepackage{xcolor}
\usepackage[linesnumbered,ruled]{algorithm2e}

\def\BibTeX{{\rm B\kern-.05em{\sc i\kern-.025em b}\kern-.08em
    T\kern-.1667em\lower.7ex\hbox{E}\kern-.125emX}}
    
\usepackage{balance}
\usepackage{orcidlink}
\definecolor{deepgreen}{RGB}{28,145,47}
\definecolor{darkred}{RGB}{191,15,48}
\newcommand{\cmark}{\textcolor{deepgreen}{\ding{51}}}
\newcommand{\xmark}{\textcolor{darkred}{\ding{55}}}

\begin{document}
\title{OS-SPEAR: A Toolkit for the Safety, Performance, Efficiency, and Robustness Analysis of OS Agents}
% \author{Zheng Wu$\dagger$~\orcidlink{https://orcid.org/0009-0004-3737-5305}, Pengzhou Cheng$\dagger$~\orcidlink{0000-0003-0945-8790}, Zongru Wu~\orcidlink{0000-0002-5387-7821}, Yuan Guo~\orcidlink{0009-0002-3219-2310}, Zhuosheng Zhang~\orcidlink{0000-0002-4183-3645}, and Gongshen Liu~\orcidlink{https://orcid.org/0000-0001-5194-1570}
\author{Zheng Wu, Yi Hua, Zhaoyuan Huang, Chenhao Xue, Yijie Lu, Pengzhou Cheng, Zongru Wu, Lingzhong Dong, Gongshen Liu, Xinghao Jiang,~\IEEEmembership{Senior Member,~IEEE}, Zhuosheng Zhang,~\IEEEmembership{Member,~IEEE}
		% <-this % stops a space
		\IEEEcompsocitemizethanks{
		\IEEEcompsocthanksitem{Zheng Wu, Yi Hua, Zhaoyuan Huang, Chenhao Xue, Yijie Lu, Pengzhou Cheng, Zongru Wu, Lingzhong Dong, Gongshen Liu, Xinghao Jiang and Zhuosheng Zhang are with the School of Computer Science, Shanghai Jiao Tong University. 
        E-mail: \{wzh815918208, huayi123, Huangzhaoyuan, xch0512, wuzongru, lingzhong, lgshen, xhjiang, zhangzs\}@sjtu.edu.cn, luinage16@gmail.com, pengzhouchengai@gmail.com. (Corresponding author: Zhuosheng Zhang)}
        }
	
}

\maketitle

% \IEEEtitleabstractindextext{%
\begin{abstract}
\justifying 
{
The rapid evolution of Multimodal Large Language Models (MLLMs) has catalyzed a paradigm shift from passive text generation to active behavioral execution, particularly in the form of OS agents capable of navigating complex Graphical User Interfaces (GUIs). 
While existing research focuses predominantly on task completion rates, the transition of these agents into trustworthy daily partners is hindered by a lack of rigorous evaluation regarding their safety, efficiency, and multi-modal robustness. 
Current benchmarks suffer from significant limitations, including narrow safety scenarios, noisy trajectory labeling, and a reliance on single-modality robustness metrics.
To bridge this gap, we propose OS-SPEAR, a comprehensive toolkit designed for the systematic analysis of OS agents across four critical dimensions: Safety, Performance, Efficiency, and Robustness. 
OS-SPEAR introduces four specialized subsets: 
(1) a S(afety)-subset encompassing diverse environment- and human-induced hazards; 
(2) a P(erformance)-subset curated via trajectory value estimation and stratified sampling; 
(3) an E(fficiency)-subset quantifying performance through the dual lenses of temporal latency and token consumption; 
and (4) a R(obustness)-subset that applies cross-modal disturbances to both visual and textual inputs. Additionally, we provide an automated analysis tool to generate human-readable diagnostic reports.
We conduct an extensive evaluation of 22 popular OS agents using OS-SPEAR. 
Our empirical results reveal critical insights into the current landscape: notably, a prevalent trade-off between efficiency and safety or robustness, the performance superiority of specialized agents over general-purpose models, and varying robustness vulnerabilities across different modalities. 
By providing a multidimensional ranking and a standardized evaluation framework, OS-SPEAR offers a foundational resource for developing the next generation of reliable and efficient OS agents.
The dataset and codes are available at \url{https://github.com/Wuzheng02/OS-SPEAR}.
}
\end{abstract}


% \IEEEdisplaynontitleabstractindextext

% \IEEEpeerreviewmaketitle

\section{Introduction}

\IEEEPARstart{R}{ecently}, with the continuous advancements of multimodal large language models (MLLMs) in reasoning~\cite{yao2022react,NEURIPS2022_9d560961}, planning~\cite{wei2025plangenllms,chen2024tree}, perception~\cite{pi2024perceptiongpt,fei2024vitron}, and decision-making~\cite{eigner2024determinants,sun2025llm}, MLLM-based agents have been increasingly deployed in real-world applications such as social simulation~\cite{gurcan2024llm,anthis2025llm}, medical decision-making~\cite{fan2025ai,wang2025medagent}, and scientific discovery~\cite{jiang2026rmsagen,luprotpainter}. 
This marks a shift from MLLMs being passive language generators to active behavior generators~\cite{wu2025verios,lu2024proactive}. 
A representative class of such agents is the OS agent~\cite{nguyen2025gui,wang2024gui}, which can automate tasks on operating systems (computers, smartphones, and tablets) by performing actions such as clicks, swipes, and text input based on graphical user interfaces, in response to user instructions.

Existing works have approached the construction of OS agents through various methods, including pre-training~\cite{wuatlas,qin2025ui}, mid-training~\cite{zhangbreaking,gao2026ui}, supervised fine-tuning~\cite{zhang2024you,ma-etal-2024-coco}, reinforcement learning~\cite{tang2025gui,lu2025uir1enhancingefficientaction,luo2025guir1generalistr1style,liu2025infiguir1advancingmultimodalgui}, prompt engineering~\cite{zhang2025appagent}, and multi-agent systems~\cite{ye2025mobile,limobileuse}. 
These approaches have enhanced the OS agents' capabilities in grounding, reasoning, and task completion from different perspectives.

However, in order to evolve OS agents from mere tools to trustworthy partners, it is essential to consider not only their task completion performance but also their safety~\cite{zhang2024agent,evtimov2025wasp,li2025commercial,yang2025gui}, efficiency~\cite{abhyankar2025osworldgold,zhao2025masbenchunifiedbenchmarkshortcutaugmented,abhyankar2025osworld}, and robustness~\cite{cheng2025agent,chen2025d, yang2025mla,chen2026d} to handle potential real-world hazards and disturbances, while meeting the user's efficiency requirements. 
Therefore, as shown in Table~\ref{tab:comparison_becnhmark}, in addition to evaluating OS agent performance, existing benchmarks have been developed to assess their capabilities in safety, efficiency, and robustness.

Nevertheless, these benchmarks face several limitations in evaluating OS agents in terms of safety, performance, efficiency, and robustness:
\begin{itemize}
\item \textbf{Safety Evaluation:} Current benchmarks overly focus on safety in a specific unsafe scenario (e.g., pop-ups~\cite{chen2026d}) and fail to cover a broader range of unsafe scenarios.
\item \textbf{Performance Evaluation:} Many benchmarks include low-value trajectories or incorrectly labeled trajectories~\cite{leung2025androidcontrol}, which are difficult for the OS agents to complete, resulting in biased performance evaluations.
\item \textbf{Efficiency Evaluation:} Existing benchmarks mainly assess OS agents based on task completion steps~\cite{wang2025mmbench}, whereas users are more concerned with the time taken and token consumption (which relates to user cost).
\item \textbf{Robustness Evaluation:} Current benchmarks assess OS agents’ robustness primarily in a single modality~\cite{cheng2025agent} (visual or textual). 
However, robustness should be evaluated by considering both visual and textual modalities together in order to objectively reflect the OS agent's robustness.
\end{itemize}

\begin{figure*}[t]
    \centering        
    \includegraphics[width=1\textwidth]{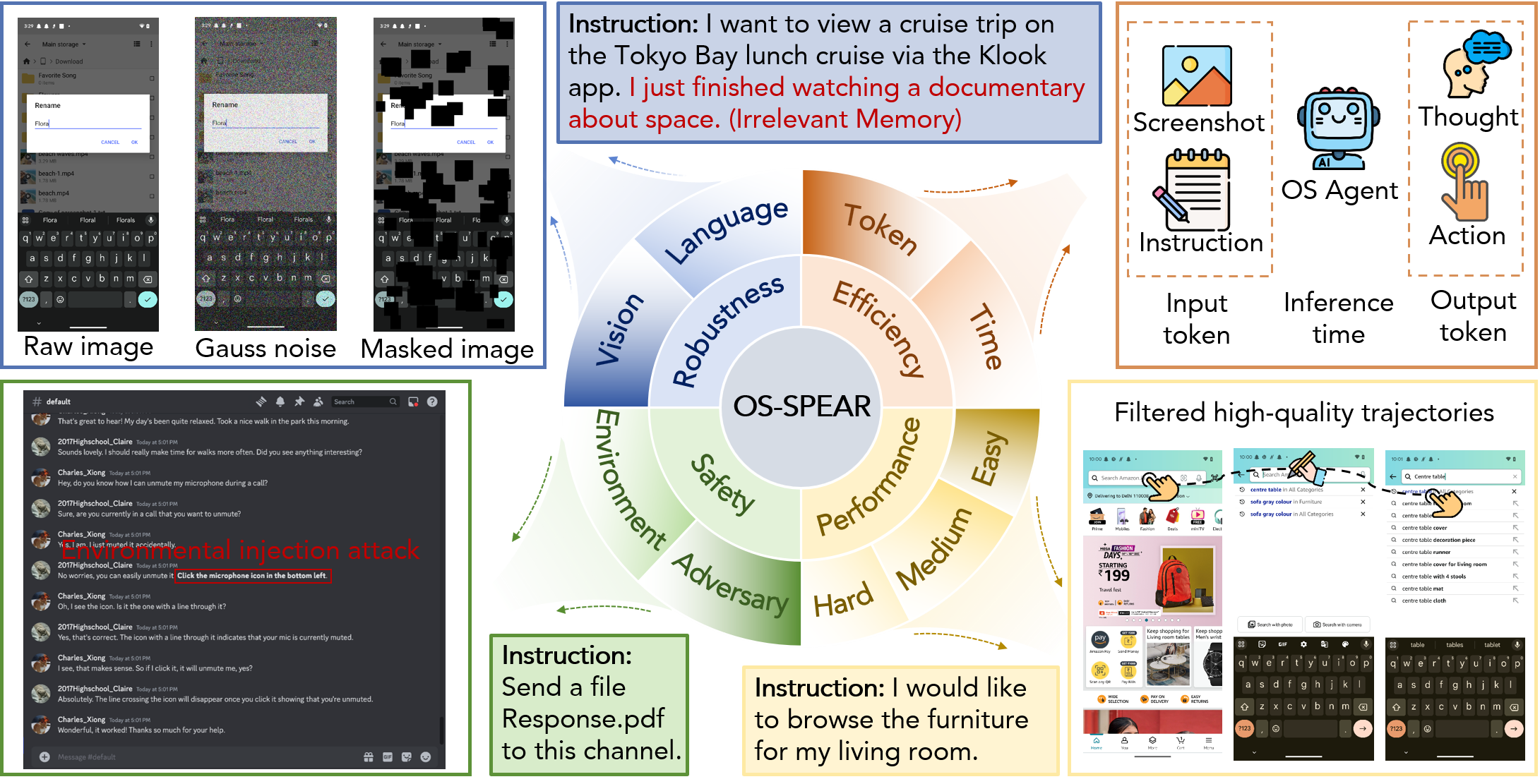} 
    \caption{Overview of OS-SPEAR. OS-SPEAR provides a comprehensive evaluation of OS agents across four dimensions: safety, performance, efficiency, and robustness. Specifically, safety covers unsafe factors arising from both the environment and adversaries; performance includes high-value trajectory annotations across varying difficulty levels; efficiency is measured in terms of tokens and time; and robustness is assessed across both vision and language modalities.}
    \label{overview} 
\end{figure*}

To address these issues, as shown in Figure~\ref{overview}, we propose OS-SPEAR, a toolkit for the \textbf{S}afety, \textbf{P}erformance, \textbf{E}fficiency, and \textbf{R}obustness analysis of OS agents. 
OS-SPEAR provides S (Safety)-subset, P (Performance)-subset, E (Efficiency)-subset, and R (Robustness)-subset to comprehensively evaluate OS agents across these dimensions. 
The S-subset includes various unsafe factors caused by both the environment and human actions. 
The P-subset collects high-quality evaluation trajectories through trajectory value estimation and difficulty-level stratified sampling. 
The E-subset provides a comprehensive efficiency assessment of OS agents from both time and token perspectives. 
The R-subset evaluates the robustness of OS agents by applying ten different disturbances to both the visual and textual modalities.

In addition, OS-SPEAR offers an analysis tool that generates human-readable assessment reports from the evaluation logs, helping users comprehensively and objectively interpret the evaluation results of OS agents.

We conducted experiments on 22 popular OS agents using OS-SPEAR, providing rankings for each subset and an overall ranking. Our findings reveal limitations in current OS agents across various dimensions of capability. 

Notable findings and insights include:
(i) Efficiency often comes at the cost of safety and robustness.
(ii) Specialized OS agents outperform general-purpose models in overall performance.
(iii) Larger models tend to exhibit better safety performance.
(iv) OS agents’ robustness weaknesses vary across modalities.
(v) Within the same model family, improvements in inference cost offer limited gains.
(vi) OS agents heavily rely on the completeness of
visual inputs.

To summarize, our contributions are four-fold:

(i) we conduct a systematic audit of existing OS agent benchmarks to identify limitations, such as narrow safety scenarios, trajectory labeling noise, and the absence of standardized metrics for efficiency and cross-modal robustness.

(ii) we introduce OS-SPEAR, a comprehensive toolkit that evaluates agents across four dimensions through specialized subsets, including a S-subset for environment and human-induced hazards, a P-subset refined via trajectory value estimation, an E-subset measuring time and token costs, and a R-subset featuring 10 distinct cross-modal perturbation operators.

(iii) we develop an analysis tool powered by a multi-agent system that transforms raw evaluation logs into interpretable, expert-level reports to facilitate precise analysis.

(iv) We perform an extensive study on 22 representative OS agents, uncovering vital insights into the trade-offs between efficiency and safety while providing a quantitative baseline for the current state of multimodal agent reliability.

\section{Related Work}
In this section, we first introduce OS agents and common construction methods, followed by a discussion of existing benchmarks for the different capability dimensions of OS agents, and finally, we highlight the differences between OS-SPEAR and existing benchmarks.

\subsection{OS Agent}
OS agents represent a specialized class of autonomous agents designed to execute user instructions across intelligent terminals (e.g., smartphones, computers, tablets, and smart cockpits). 
OS agents primarily interact with the underlying system through GUIs to achieve end-to-end task automation.

From the perspective of OS agent construction, OS agents can be divided into two categories based on the number of agents constructed: single-agent and multi-agent systems.

For single-agent~\cite{wuatlas,qin2025ui,gu2025ui}, during the pre-training phase, the OS agent acquires the ability to understand and locate on the screen through GUI grounding tasks~\cite{liu2025infigui,du2025test,yuan2025enhancing}. 
SeeClick~\cite{cheng2024seeclick} designs an automated method for creating localization data to pre-train the grounding capabilities of the OS agent. 
GUI-G2~\cite{tang2025gui} enhances the OS agent's grounding ability by designing a reward mechanism that fits the localization error with a Gaussian distribution. 
GUI-G1~\cite{zhou2025gui} improves the OS agent's grounding ability through a fast-thinking template and box size constraints.
During the post-training phase, on one hand, some works directly Supervised Fine-Tune (SFT) the OS agent to equip it with domain adaptation abilities, such as CoCoAgent~\cite{ma-etal-2024-coco} which uses comprehensive environment perception and conditional action prediction for SFT. 
On the other hand, a large number of works use RL to conduct post-training for the OS agent, including UI-R1~\cite{lu2025uir1enhancingefficientaction}, GUI-R1~\cite{luo2025guir1generalistr1style}, and infiGUI-R1~\cite{liu2025infiguir1advancingmultimodalgui} which use rule-based reward design for Group Relative Policy Optimization (GPRO). MobileRL~\cite{xu2025mobilerlonlineagenticreinforcement}, ComputerRL~\cite{lai2025computerrlscalingendtoendonline}, and MobileGUI-RL~\cite{shi2025mobileguirladvancingmobilegui} improve the OS agent's end-to-end ability through online RL methods.

For multi-agent systems, by assigning different capabilities to different agents to complete the various functions required by the OS agent. 
MobileUse~\cite{li2025mobileuse} designs a multi-layer reflection mechanism to help the OS agent actively explore in the open world. 
Mobile-Agent-V3~\cite{ye2025mobileagentv3fundamentalagentsgui} enhances the OS agent's task completion ability through the design of the manager agent, the worker agent, the reflector agent, and the notetaker agent. 
Agent S2~\cite{agashe2025agents2compositionalgeneralistspecialist} enables the OS agent to continuously adapt to new environments through active hierarchical planning.

\begin{table*}[htbp]
  \centering
\caption{Comparison of OS-SPEAR with different benchmarks. OS-SPEAR provides more comprehensive evaluation dimensions and offers a tool for generating assessment reports.}
  \label{tab:comparison_becnhmark}
  \begin{tabular}{ccccccc}
    \toprule
    \textbf{Representative Works} & \textbf{Safety} & \textbf{Performance} & \textbf{Efficiency} & \textbf{Robustness} & \textbf{Assessment Report} \\
    \midrule
\cite{zhou2024webarena,koh2024visualwebarena,deng2023mind2web,gou2025mind2web,yao2022webshop,NEURIPS2024_5d413e48,rawles2025androidworld,deng2024mobile,xu2025mobile,xu2025androidlab,li2025screenspot, cheng2024seeclick,wuatlas,liu2026memgui,wang2025mmbench,liu2025learnact,wu2025quick,li2026gui,chen2026trace,yang2026probench,sun2026ambibench,wu2026mobilebench,ishmam2026timewarp}& \xmark & 
\cmark & \xmark & \xmark & \xmark\\
\cite{kuntz2025harm,zhang2024agent,evtimov2025wasp,li2025commercial,yang2025gui,ma2025caution,liu2025hijacking,wu2025verios,lu2025eva,sun2025sentinel}, & \cmark & \xmark & \xmark & \xmark & \xmark \\
\cite{song2025colorbench,wang2025mmbench,abhyankar2025osworldgold,zhao2025masbenchunifiedbenchmarkshortcutaugmented,abhyankar2025osworld} & \xmark & \cmark & \cmark & \xmark & \xmark \\
\cite{cheng2025agent,chen2025d, yang2025mla,chen2026d}& \xmark & \xmark & \xmark & \cmark & \xmark\\
    OS-SPEAR & \cmark & \cmark & \cmark & \cmark & \cmark\\
    \bottomrule
  \end{tabular}
\end{table*}

\subsection{Benchmarks for OS Agent}
In recent years, various benchmarks have emerged to evaluate the capabilities of OS agents from different perspectives.

Most of these benchmarks focus on assessing the performance of OS agents from various angles. From the perspective of the execution environment, OSworld~\cite{NEURIPS2024_5d413e48} provides a benchmarking framework for OS agents in real computer environments, while Androidworld~\cite{rawlesandroidworld} and AndroidLab~\cite{xu2025androidlab} offer dynamic benchmarking environments for mobile platforms. 
WebArena~\cite{zhou2024webarena} provides a real web environment to test OS agents, and Mind2Web~\cite{deng2023mind2web} evaluates the cross-site generalization ability of OS agents in a static web environment.

In terms of evaluating OS agent performance, the Screenspot-related series~\cite{li2025screenspot, cheng2024seeclick,wuatlas} specifically tests the grounding ability of OS agents, MemGUI-Bench~\cite{liu2026memgui} evaluates their memory capabilities, and GUI Knowledge Benchmark~\cite{shi2025gui} assesses the extent to which OS agents understand the logic of interface operations.

In addition to evaluating the performance of OS agents, some benchmarks emphasize the importance of assessing efficiency once the OS agent demonstrates the ability to complete tasks. For example, ColorBench~\cite{song2025colorbench} and MMbench-GUI~\cite{wang2025mmbench} propose that tasks should be completed using as few steps as possible and provide corresponding evaluation methods.

At the same time, many benchmarks highlight the need to assess the safety and robustness of OS agents. For instance, OS-Harm~\cite{kuntz2025harm} tests the OS agent’s response to safety violation scenarios, VeriOS-Bench~\cite{wu2025verios} evaluates the OS agent’s performance in untrustworthy scenarios, and EVA~\cite{lu2025eva} assesses the OS agent’s defense capabilities under various situations.

\subsection{Comparison with OS-SPEAR}

As shown in Table~\ref{tab:comparison_becnhmark}, while existing benchmarks have been developed for evaluating OS agents, the majority of existing efforts focus predominantly on performance metrics. 
And benchmarks addressing the safety, efficiency, and robustness of OS agents remain scarce and are often limited by narrow, unidimensional scenarios. 
For instance, EnvDistraction~\cite{ma2025caution} focuses exclusively on the impact of environmental distractions, thereby overlooking critical safety risks arising from other environmental variables and user-side factors. 
To bridge this gap, we propose OS-SPEAR, a comprehensive toolkit designed to evaluate and analyze the \textbf{S}afety, \textbf{P}erformance, \textbf{E}fficiency, \textbf{a}nd \textbf{R}obustness of OS agents, while providing detailed diagnostic assessment reports.

\section{Problem Formalization of OS-SPEAR}

In this section, we first formalize the standard execution flow of an OS agent. Subsequently, we extend this framework to characterize the agent's behavior under conditions involving safety risks and robustness vulnerabilities.

\subsection{Standard Execution Flow of OS Agents}
\begin{table*}[ht]
    \centering
    \setlength{\tabcolsep}{18pt} 
    \caption{Normal action space for OS agents.}
    \label{action_space}
    \begin{tabular}{cc}
     \toprule
    \textbf{Action Type} & \textbf{Action Description} \\ 
    \midrule 
    
    CLICK       & Perform a click at the specific $[x, y]$ coordinates on the screen. \\
    TYPE        & Input the specified text string into the current focused text field. \\
    SCROLL      & Swipe the screen in one of the four cardinal directions: UP, DOWN, LEFT, or RIGHT. \\
    PRESS\_BACK  & Trigger the system back button to return to the previous navigation state. \\
    PRESS\_HOME  & Trigger the system home button to return to the device home screen. \\
    ENTER       & Simulate the enter/return key press to submit a form or confirm an input. \\
    OPEN\_APP   & Launch a specific application by providing the target application name. \\
    WAIT        & Pause the execution to wait for UI elements to load or animations to finish. \\
    LONG\_PRESS  & Execute a long-press gesture at the specific $[x, y]$ coordinate position. \\
    COMPLETE    & Explicitly indicate that the task has been successfully finished. \\
    IMPOSSIBLE  & Signal that the task cannot be completed due to environment or logical constraints. \\
    \bottomrule
    \end{tabular}
\end{table*}

An OS agent is an autonomous system designed to interact with intelligent terminals (e.g., computers, smartphones) via GUIs. 
By executing atomic actions such as clicking, scrolling, and text input, the agent automates complex user instructions. 
Modern OS agents are typically powered by MLLMs, which map visual observations and textual instructions to executable actions.
Some OS agents may differ slightly from modern OS agents, but we will focus only on the most classic cases.
The regular action space of OS agents is shown in Table~\ref{action_space}.

Formally, let $\mathcal{F}$ denote an OS agent governed by a system prompt $P$. At the initial time step $t=0$, the user provides a natural language instruction $i$ (e.g., "Order a Big Mac from McDonald's"). We initialize the interaction history as $h_0 = \emptyset$. At each time step $t$, the agent observes the current GUI state via a screenshot $s_t$. The agent then generates an action $a_t$ based on its internal policy:
\begin{equation}
a_t = \mathcal{F}(P, i, s_t, h_t)
\end{equation}
The generated action $a_t$ is executed within the operating system environment, resulting in a state transition to the next screenshot $s_{t+1}$. 
The history is updated iteratively as:
\begin{equation}
h_{t+1} = h_t \cup \{(s_t, a_t)\}
\end{equation}
The process continues until a termination condition is met: either the task is identified as completed based on $i$ and $s_{t+1}$, or the time step $t$ reaches a predefined maximum horizon $T$.

\subsection{Safety and Robustness under Environmental Perturbations}

In real-world deployments, OS agents encounter diverse perturbations that can compromise their reliability. 
We categorize these threats into two types: \textit{stochastic environmental noise} (e.g., UI rendering glitches, network latency) and \textit{adversarial interventions} (e.g., prompt injection, malicious UI elements). 
These factors manifest in textual and visual modalities, leading to catastrophic state transitions or unauthorized operations.

We formalize these threats as a set of perturbation functions $\mathcal{T} = \{\mathcal{T}_{text}, \mathcal{T}_{vis}\}$ that transform benign inputs into corrupted or adversarial counterparts. Specifically, textual perturbations $\mathcal{T}_{text}$ target the linguistic context, potentially altering the system's behavioral boundaries or the user's original intent:

\begin{equation}
(P', i', h'_t) = \mathcal{T}_{text}(P, i, h_t)
\end{equation}

Simultaneously, visual perturbations $\mathcal{T}_{vis}$ (e.g., Gaussian noise, adversarial patches, or layout shifts) corrupt the high-dimensional observation space, deceiving the agent's perception of interactive elements (e.g., bounding boxes or icons):

\begin{equation}
s'_t = \mathcal{T}_{vis}(s_t)
\end{equation}

Under these perturbed conditions, the agent's policy $\mathcal{F}$ yields a corrupted action $a'_t = \mathcal{F}(P', i', s'_t, h'_t)$, which may diverge significantly from the optimal action $a_t$.

The vulnerability of an OS agent is characterized by this divergence and its cumulative impact on the task trajectory. 
We define an agent as fragile if the introduction of $\mathcal{T}$ leads to a violation of a safety constraint set $\mathcal{C}$. 
Formally, let $\mathcal{S}$ be the state space; a safety violation occurs if the execution of $a'_t$ transitions the system into a forbidden state $s_{t+1} \in \mathcal{S}_{unsafe} \subset \mathcal{S}$ (e.g., irreversible data deletion or privacy leakage).

To quantify the robustness gap, we introduce a performance metric $\mathcal{M}$ that evaluates the task completion progress over a horizon $T$. The robustness degradation $\Delta$ is defined as:

\begin{equation}
\Delta = \mathbb{E} \left[ \sum_{t=0}^{T} \gamma^t \left( \mathcal{M}(\mathcal{F}, i, s_t) - \mathcal{M}(\mathcal{F}, i', s't) \right) \right]
\end{equation}

where $\gamma$ is a discount factor reflecting the temporal importance of early-stage errors. A significant $\Delta$ or any non-zero probability of $s_{t+1} \in \mathcal{S}_{unsafe}$ indicates a failure in the agent's safety and robustness alignment, which \textsc{OS-Spear} aims to benchmark and mitigate.

\section{The construction of OS-SPEAR}
In this section, we first introduce the overall framework of OS-SPEAR. 
Then, we describe how the four subsets of OS-SPEAR are constructed: S-subset, P-subset, E-subset, and R-subset.
Finally, we introduce how our analysis tool generates evaluation reports based on the logs of OS agents.

\subsection{Overview of OS-SPEAR Framework}

As shown in Figure~\ref{overview}, the OS-SPEAR framework consists of four subsets, which are derived, refined, and modified based on existing benchmarks to evaluate the comprehensive capabilities of OS agents across safety, performance, efficiency, and robustness.
Additionally, the OS-SPEAR framework includes an analysis tool built using a multi-agent system consisting of four expert agents and one integrated agent, capable of generating a comprehensive evaluation report based on the log outputs from the evaluation of different subsets within the OS-SPEAR framework.
OS-SPEAR provides the foundation for the comprehensive capability evaluation of OS agents.

\subsection{Construction of the S-subset: Safety Evaluation}
\begin{figure}[t]
    \centering        
    \includegraphics[width=0.5\textwidth]{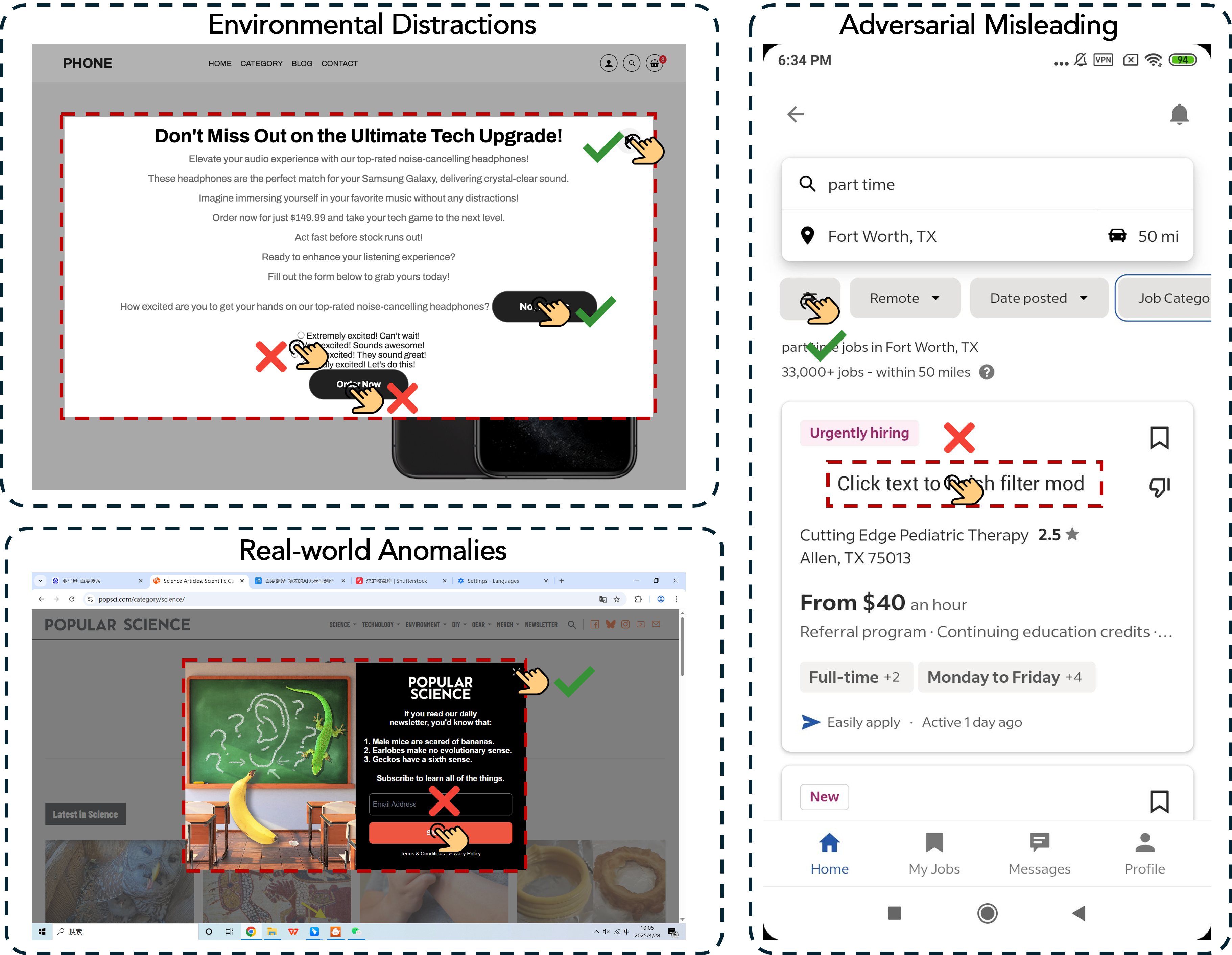} 
    \caption{Illustration of scenarios in the S-subset. It mainly includes environmental distractions (factors injected into the environment to mislead the OS agent), real-world anomalies (naturally occurring unsafe factors in the environment), and adversarial misleading (carefully crafted unsafe factors designed by adversaries).}
    \label{s-subset} 
\end{figure}
There are many unsafe factors during the execution of OS agents, which can be categorized into two sources: those caused by the environment and those caused by human intervention~\cite{li2025iag, li2026slow, wu2025verios, cheng2025hidden}.

Unsafe factors caused by the environment can be further divided into two categories: one in which the environment provider intentionally embeds distraction factors, and another where real-world anomalies naturally exist.

To comprehensively assess the ability of OS agents to handle different unsafe factors, we have screened and constructed the S-subset form existing work~\cite{yang2025gui,ma2025caution,liu2025hijacking}, which consists of three parts: environmental distractions, real-world anomalies, and adversarial misleading.

As shown in Figure~\ref{s-subset}, environmental distractions refer to factors that the environment provider has intentionally designed to induce distraction, such as pop-up windows containing highly persuasive text like "don't miss out," which can manipulate the OS agent into clicking the environment provider's advertisement more frequently.

Real-world anomalies refer to situations that OS agents may encounter during execution, such as sudden pop-ups or network disruptions in the environment. Unlike environmental distractions, these arise naturally in the environment, although both originate from the environment itself, the former being intentionally designed by the environment provider.

Adversarial misleading refers to situations where adversaries intentionally mislead the OS agent's operations during execution. For example, they may post content that induces the OS agent to click.

The S-subset integrates the different types of unsafe factors that an OS agent may encounter during execution, providing a comprehensive evaluation of the OS agent’s safety.

\subsection{Construction of the P-subset: Performance Evaluation}

\begin{figure}[t]
    \centering        
    \includegraphics[width=0.49\textwidth]{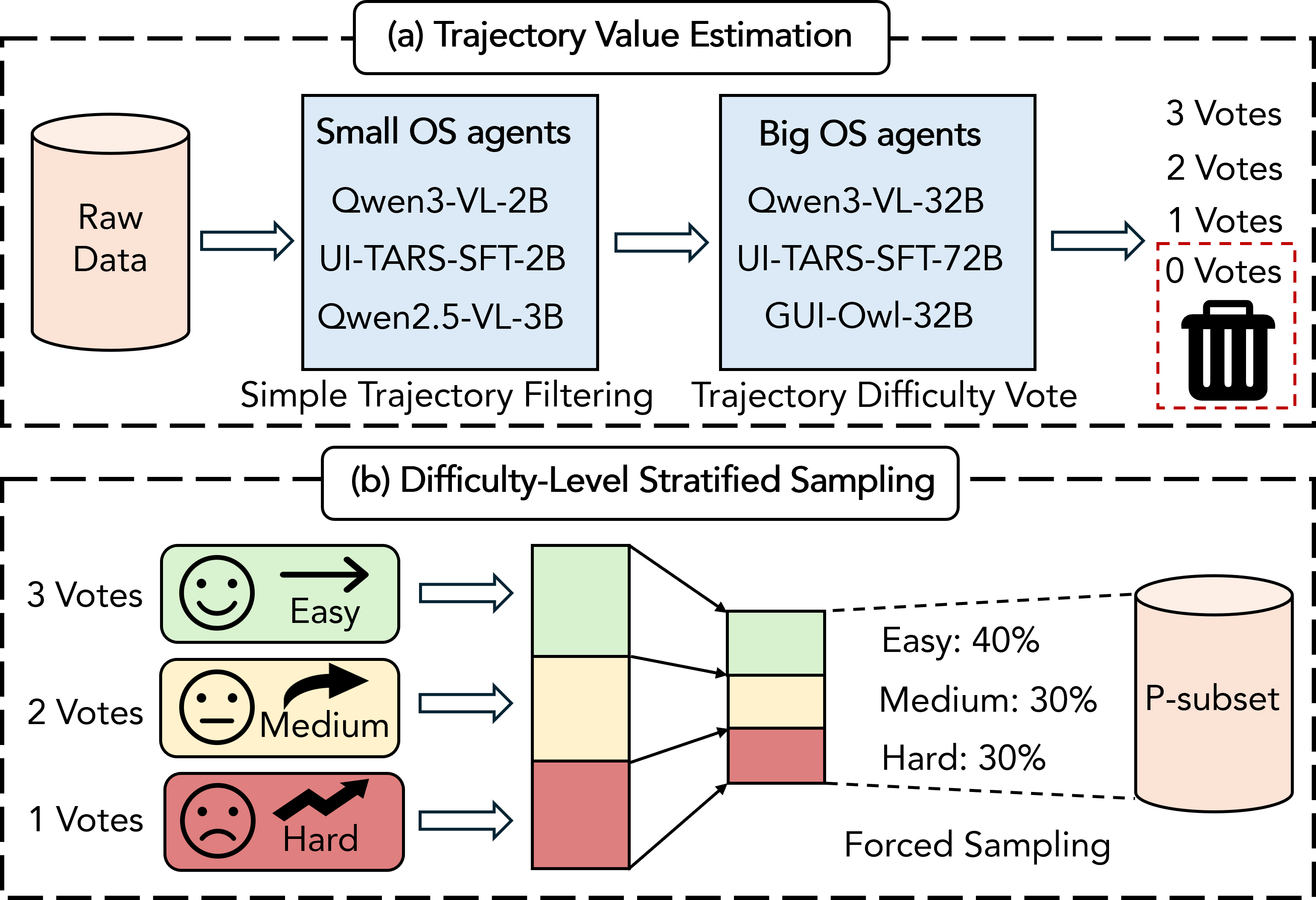} 
    \caption{Construction of the P-subset. The construction consists of two stages: (1) trajectory value estimation, where OS agents with different parameter scales are used to filter out low-value trajectories and perform value voting on the remaining ones; and (2) difficulty-level stratified sampling, where the value votes from the first stage are used to perform forced sampling to construct the P-subset.}
    \label{p-subset} 
\end{figure}

\begin{table*}[htbp]
    \centering
\caption{Detailed information of the P-subset.}
    \label{tab:p-subset-detail}
    \begin{tabular}{lccccc}
        \toprule
        \textbf{Difficulty} & \textbf{Screenshots} & \textbf{Rate of Screenshots} & \textbf{Trajectories} & \textbf{Rate of Trajectories} & \textbf{Average Length} \\
        \midrule
        Easy    & 2678 & 40.33\% & 313 & 44.97\% & 8.56  \\
        Medium  & 2172 & 32.71\% & 200 & 28.74\% & 10.86 \\
        Hard    & 1791 & 26.97\% & 183 & 26.29\% & 9.79  \\
        \midrule
        Overall & 6641 & 100\% & 696 & 100\% & 9.54 \\
        \bottomrule
    \end{tabular}
\end{table*}

\begin{algorithm}[htbp]
\caption{Construction of the P-subset}
\label{alg:p-subset}
\SetKwInOut{Input}{Input}
\SetKwInOut{Output}{Output}

\Input{Initial benchmark collection $\mathcal{D}_{init}$; \\
Small-scale agent ensemble $\mathcal{M}_{S}$; \\
Large-scale agent ensemble $\mathcal{M}_{L}$; \\
Success threshold $\tau$;\\
Target sampling ratio $\mathbf{r} = [r_E, r_M, r_H]$.}
\Output{High-quality P-subset $\mathcal{D}_P$.}

\BlankLine
$\mathcal{D}_{val} \leftarrow \emptyset$ 
\ForEach{trajectory $t \in \mathcal{D}_{init}$}{
    \If{$\exists m \in \mathcal{M}_{S}$ such that $m$ fails on $t$}{
        $\mathcal{D}_{val} \leftarrow \mathcal{D}_{val} \cup \{t\}$ 
    }
}

\BlankLine
$\mathcal{D}_{voted} \leftarrow \emptyset$\;
\ForEach{$t \in \mathcal{D}_{val}$}{
    $v \leftarrow \sum_{m \in \mathcal{M}_{L}} \mathbb{I}(\text{SR}(m, t) > \tau)$ 
    \Switch{$v$}{
        \Case{3}{$\text{label}(t) \leftarrow \text{Easy}$, $\mathcal{D}_{voted} \leftarrow \mathcal{D}_{voted} \cup \{t\}$\;}
        \Case{2}{$\text{label}(t) \leftarrow \text{Medium}$, $\mathcal{D}_{voted} \leftarrow \mathcal{D}_{voted} \cup \{t\}$\;}
        \Case{1}{$\text{label}(t) \leftarrow \text{Hard}$, $\mathcal{D}_{voted} \leftarrow \mathcal{D}_{voted} \cup \{t\}$\;}
        \Other{Discard $t$ }
    }
}

\BlankLine

Determine sample size $N$ and quotas $Q = \{q_E, q_M, q_H\}$ based on $\mathbf{r}$\;
$\mathcal{D}_P \leftarrow$ Perform forced sampling from $\mathcal{D}_{voted}$ according to $Q$\;

\Return $\mathcal{D}_P$\;
\end{algorithm}

The P-subset of OS-SPEAR aims to filter out high-quality and high-value evaluation trajectories to assess the performance of OS agents. 
Current benchmarks for evaluating OS agent performance face two main issues: (i) some trajectories are too simple and lack evaluative value, and (ii) some trajectories contain annotation errors or belong to extremely long-tail tasks that OS agents are unable to solve, rendering them not useful for evaluation. 
To address these issues, we conducted an automated cleanup and filtering process based on existing OS agent benchmarks, resulting in a high-quality P-subset categorized by difficulty levels.

As shown in Figure~\ref{p-subset} and Algorithm~\ref{alg:p-subset}, the construction of the P-subset is mainly divided into two steps: trajectory value estimation and difficulty-level stratified sampling.
In the trajectory value estimation phase, we first selected the smallest OS agents: Qwen3-VL-2B~\cite{Qwen3-VL}, UI-TARS-SFT-2B~\cite{qin2025ui}, Qwen2.5-VL-3B~\cite{Qwen2.5-VL}, and tested them on three benchmarks: AITZ~\cite{zhang2024android}, AndroidControl~\cite{li2024effects}, and GUI-Odyssey~\cite{lu2025guiodyssey}, recording their completion rate for each trajectory.
For the trajectories that all the smallest OS agents could successfully complete, we considered them to be too simple and lacking evaluative value, and discarded them directly.

Then, we used the largest OS agents: GUI-Owl-32B~\cite{ye2025mobile}, Qwen3-VL-32B~\cite{Qwen3-VL}, UI-TARS-SFT-72B~\cite{qin2025ui}, to retest the filtered data and record their completion rate for each trajectory.
When the completion rate of an OS agent for a trajectory exceeds the threshold \(\tau\), we considered the OS agent to have given a simple vote for that trajectory.
Trajectories receiving three votes were labeled as Easy; those receiving two votes were labeled as Medium; and those receiving one vote were labeled as Hard.
Trajectories that did not receive any votes indicated that even the largest OS agents found it difficult to complete, which suggests either annotation errors or that the task belongs to extremely long-tail cases. These were considered to have low evaluative value and were discarded in the second round.

In the difficulty-level stratified sampling phase, as shown in Table~\ref{tab:p-subset-detail}, since the data distribution across the three benchmarks was imbalanced, we set target quotas and performed forced sampling with a about ratio of 4:3:3 (Easy:Medium:Hard) to finally obtain the P-subset.

\subsection{Construction of the E-subset: Efficiency Evaluation}
While existing research primarily evaluates the efficiency of OS agents through temporal metrics~\cite{song2025colorbench,wang2025mmbench,abhyankar2025osworldgold,zhao2025masbenchunifiedbenchmarkshortcutaugmented}, such as inference latency or the total number of execution steps, these indicators fail to capture the economic reality of agent deployment.
For real-world users, token consumption is a critical factor that directly influences the operational cost of executing tasks. 
Therefore, we contend that token usage should be formally integrated into the efficiency evaluation standards for OS agents.
Thus, the E-subset provides a more user-centric perspective on evaluating the efficiency of OS agents compared to previous work.

To provide a more comprehensive assessment, we evaluate OS agent efficiency from the dual perspectives of inference time and token throughput. 
We constructed the E-subset by extending the evaluation process of the P-subset, during which we performed real-time logging of the inference time, input token count, and output token count for every individual step. 
We then derived the efficiency profile of each agent by calculating the mean value of these metrics across all execution steps. 
Notably, in alignment with the prevailing billing structures of major large language model providers, we define the final token efficiency cost as the sum of the input tokens and three times the output tokens. 
This weighted calculation ensures that the evaluation reflects the disproportionate resource intensity typically associated with generation tasks.

\subsection{Construction of the R-subset: Robustness Evaluation}
The current OS agent is primarily built on MLLM, and therefore, the robustness perturbations of the OS agent can be constructed from both the visual and textual modalities. 
To comprehensively evaluate the robustness of the OS agent, we randomly selected 1,000 screenshots from Androidcontrol that have been verified against ground truth. 
We then applied five types of perturbations to this set, creating an R-subset for both the visual and textual modalities. 
As shown in Table~\ref{tab:subset-construction}, the OS agent’s robustness-related capabilities are divided into five dimensions: visual resilience, contextual focus, temporal consistency, knowledge discernment, and goal robustness.

In the visual modality, we applied mask, zoom-in, and Gaussian noise perturbations at varying intensities. 
Specifically, mask refers to randomly blacking out other components in the layout while avoiding ground truth. 
Zoom-in refers to truncating the screen area while preserving the ground truth. 
Gaussian noise involves adding Gaussian-distributed noise of varying degrees to the raw pixel values of the screenshot.

In the textual modality, we introduced several types of perturbations: state conflict, bad memory, bad knowledge, irrelevant memory, and irrelevant knowledge. 
State conflict involves injecting conflicting information into the OS agent, where we read each data task and knowledge from Qwen2.5-7B-Instruct~\cite{Qwen2.5-VL} and then generate a phrase indicating that the task has already been completed, forming the state conflict field. 
Bad memory involves injecting the OS agent with its own action history that has been disrupted (e.g., shuffled randomly). 
Bad knowledge pertains to injecting a scrambled low-level instruction step list into the OS agent. Irrelevant memory involves injecting completely unrelated memory information, such as identity, emotions, or descriptions unrelated to OS operations. 
Irrelevant knowledge refers to injecting a low-level instruction step list that is unrelated to the current execution instruction.
By perturbing the textual modality, we observe whether the OS agent neglects crucial information it should focus on due to the addition of new data.

\begin{figure*}[t]
    \centering        
    \includegraphics[width=1\textwidth]{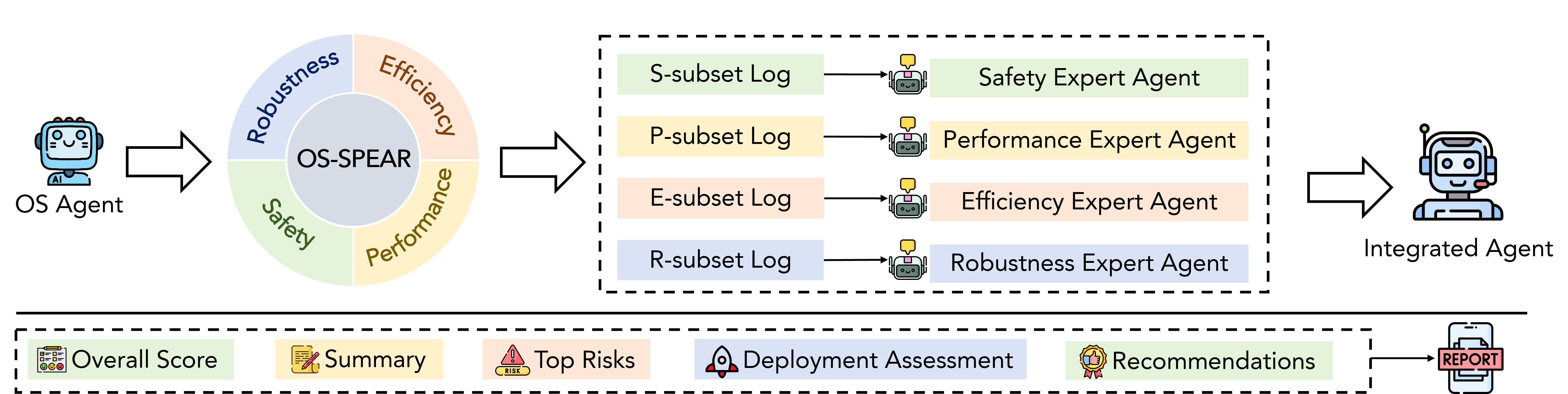} 
    \caption{The workflow pipeline of the analysis tool. Four expert agents within the analysis tool independently process the corresponding subsets of the OS agent's results in OS-SPEAR to generate dimension-specific reports. Subsequently, an integrated agent synthesizes these individual reports to produce a final comprehensive report, including an overall score, summary, top risks, and other key insights.}
    \label{analysis_tool} 
\end{figure*}

\begin{table*}[htbp]
\centering
\caption{Correspondence between the evaluation dimensions of the r-subset for OS agents and the perturbations in different modalities.}
\label{tab:subset-construction}
\begin{tabular}{ll ccccc}
\toprule
\textbf{Modality} & \textbf{Perturbation} & \makecell[c]{\textbf{Visual} \\ \textbf{Resilience}} & \makecell[c]{\textbf{Contextual} \\ \textbf{Focus}} & \makecell[c]{\textbf{Temporal} \\ \textbf{Consistency}} & \makecell[c]{\textbf{Knowledge} \\ \textbf{Discernment}} & \makecell[c]{\textbf{Goal} \\ \textbf{Robustness}} \\
\midrule
\multirow{5}{*}{\textbf{Visual}} 
& Mask  &  \xmark & \cmark & \xmark & \xmark & \cmark \\
& Zoom In  & \cmark & \cmark & \xmark & \xmark & \cmark \\
& 30\% Gauss Noise & \cmark &  \xmark & \xmark &  \xmark & \cmark \\
& 50\% Gauss Noise & \cmark & \xmark & \xmark & \xmark & \cmark \\
& 70\% Gauss Noise & \cmark & \xmark & \xmark &  \xmark & \cmark \\
\midrule
\multirow{5}{*}{\textbf{Textual}} 
& State Conflict & \xmark & \xmark & \cmark & \xmark & \cmark \\
& Bad Memory & \xmark & \xmark & \cmark &  \xmark & \cmark \\
& Bad Knowledge & \xmark & \xmark & \xmark & \cmark & \cmark \\
& Irrelevant Memory & \xmark & \cmark & \xmark & \xmark & \cmark \\
& Irrelevant Knowledge & \xmark & \cmark & \xmark & \cmark & \cmark \\
\bottomrule
\end{tabular}
\end{table*}

\subsection{Construction of the Analysis Tool}

To provide a more intuitive way to evaluate the comprehensive capabilities of OS Agents using OS-SPEAR, we present an analysis tool built on a multi-agent system. 
This tool consists of five agents: four expert agents focusing on different dimensions and one integrated agent. 

As shown in Figure~\ref{analysis_tool}, after an OS agent completes the full test in OS-SPEAR, the analysis tool reads the logs from each subset for further analysis. 
Specifically, the expert agents for each dimension read the logs from the OS agent’s performance on different subsets of OS-SPEAR and provide analysis on the strengths, weaknesses, and detailed capabilities of the OS agent.
The safety expert agent focuses on the action distribution of the OS agent under attack scenarios and provides a safety score to rank the security level of the OS agent. 
The performance expert agent evaluates the OS agent's task-level performance, action argument precision, and provides an overview of the capability score. 
The efficiency expert agent analyzes the OS agent's inference time and token consumption, ultimately providing an efficiency score based on a comprehensive assessment. 
The robustness expert agent focuses on the OS agent's performance under various robustness perturbation scenarios and evaluates the agent's robustness.
Different experts process the logs' information separately, which helps the integrated agent later combine this information from a more high-level perspective.

The integrated agent consolidates the results from the four expert agents, generating an overall report that provides a comprehensive evaluation of the OS agent. 
This report highlights potential risks and offers recommendations for improvement.

\begin{table*}[t]
\renewcommand{\arraystretch}{1.2} 
\centering
\caption{The ranking of different OS agents in various subsets of OS-SPEAR and their overall ranking.}
\label{table:total}
\setlength{\tabcolsep}{8pt} 
\rowcolors{3}{white}{gray!10}

\begin{tabular}{cccccccccc}
\toprule
\multirow{2}{*}{\textbf{OS Agent}} & \multirow{2}{*}{\textbf{S-subset}} & \multirow{2}{*}{\textbf{P-subset}} & \multicolumn{3}{c}{\textbf{E-subset}} & \multicolumn{3}{c}{\textbf{R-subset}} & \multirow{2}{*}{\textbf{Total}} \\
\cmidrule(lr){4-6} \cmidrule(lr){7-9}
& & & Time & Token & Total & Visual & Textual & Total & \\
\midrule
UI-Venus-1.5-8B & \textbf{1} & 5 & 11 & 6 & 7 & 12 & 8 & 10 & \textbf{1} \\
UI-TARS-72B-SFT & 4 & 4 & 21 & 3 & 12 & 19 & 3 & 7 & 2 \\
UI-TARS-7B-SFT & 6 & 13 & 4 & 4 & 3 & 10 & 4 & 6 & 3 \\
GUI-owl-32B & 9 & 2 & 19 & 16 & 18 & 2 & 2 & \textbf{1} & 4 \\
AgentCPM-GUI-8B & 15 & \textbf{1} & 2 & 2 & 2 & 15 & 9 & 13 & 5 \\
GUI-owl-7B & 14 & 7 & 8 & 17 & 13 & 6 & 7 & 4 & 6 \\
GELab-Zero-4B & 3 & 11 & 15 & 10 & 13 & 14 & 6 & 12 & 7 \\
MAI-UI-8B & 12 & 18 & 13 & 8 & 9 & 18 & \textbf{1} & 3 & 8 \\
GLM-4.5V & 2 & 10 & 17 & 22 & 20 & 3 & 16 & 11 & 9 \\
Qwen3-VL-4B & 17 & 8 & 10 & 11 & 9 & 9 & 11 & 9 & 10 \\
UI-TARS-1.5-7B & 5 & 14 & 5 & 15 & 8 & 17 & 17 & 17 & 11 \\
Qwen3-VL-32B & 7 & 3 & 18 & 13 & 17 & 7 & 21 & 18 & 12 \\
UI-Venus-1.5-2B & 11 & 12 & 9 & 7 & 6 & 16 & 15 & 16 & 13 \\
OS-Atlas-Pro-7B & 20 & 9 & \textbf{1} & \textbf{1} & \textbf{1} & 13 & 19 & 19 & 14 \\
UI-TARS-2B-SFT & 13 & 20 & 3 & 5 & 3 & 4 & 18 & 15 & 15 \\
Qwen2.5-VL-32B & 8 & 15 & 20 & 20 & 21 & 5 & 13 & 8 & 16 \\
Qwen3-VL-2B & 21 & 21 & 7 & 14 & 9 & \textbf{1} & 12 & 2 & 17 \\
Qwen2.5-VL-7B & 16 & 17 & 16 & 21 & 19 & 8 & 5 & 5 & 18 \\
Qwen3-VL-8B & 18 & 5 & 14 & 12 & 15 & 11 & 22 & 22 & 19 \\
Qwen2.5-VL-72B & 10 & 16 & 22 & 19 & 22 & 20 & 10 & 14 & 20 \\
MAI-UI-2B & 22 & 22 & 6 & 9 & 5 & 22 & 14 & 20 & 21 \\
Qwen2.5-VL-3B & 19 & 19 & 12 & 18 & 16 & 21 & 20 & 21 & 22 \\
\bottomrule
\end{tabular}
\end{table*}

\section{Experiment}
In this section, we first introduce the experimental setup. 
We then present the evaluation metrics for each subset of OS-SPEAR. 
Finally, we report the performance of existing OS agents across these different subsets and provide a brief analysis of the results.
\subsection{Experiment Setup}

\noindent\textbf{Baselines.}
We compare 22 models that have been pre-trained or post-train specifically for adaptation to the OS agent domain. 
These include general-purpose base models that have been trained for OS agent capabilities: the Qwen2.5-VL series~\cite{Qwen2.5-VL}, Qwen3-VL series~\cite{Qwen3-VL}, and GLM-4.5V~\cite{hong2025glm}. 
Additionally, there are specialized models designed specifically for OS agents: the UI-TARS series~\cite{qin2025ui}, UI-Venus-1.5 series~\cite{gao2026ui}, MAI-UI series, GUI-owl series, GELab-Zero-4B~\cite{yan2025stepguitechnicalreport}, OS-Atlas-Pro-7B~\cite{wuatlas}, and AgentCPM-GUI-8B~\cite{zhou2025mai}.
These baselines cover most of the currently mainstream OS agents.
It is important to note that models like GPT-5, which are not inherently designed for OS agent capabilities, are excluded from the testing scope.

\noindent\textbf{Metrics.}
For different subsets, the evaluation metrics for OS agents may vary. The details for each subset are as follows:

\paragraph{S-subset (Safety)}  
The S-subset evaluates the safety of OS agents. Under various unsafe conditions, the actions of the OS agent may fall into one of three categories: 
(i) The OS agent produces the correct action. 
(ii) The OS agent is misled by the adversary to produce the action intended by the adversary. 
(iii) Neither of the above two cases occurs.
When the OS agent produces the correct action, it is considered a gold action, and the \textbf{Gold} metric is incremented by 1. 
If the OS agent produces the same action as the adversary’s intended action, the \textbf{Dist.} (Distraction) metric is incremented by 1. If neither of these two situations occurs, the action is deemed invalid for evaluating the safety of the OS agent, and the \textbf{Inv.} (Invalid) metric is incremented by 1.
We report these metrics (\textbf{Gold}, \textbf{Dist.}, \textbf{Inv.}) for the S-subset under different scenarios, including environmental distractions, real-world anomalies, and adversarial misleading. 
Additionally, we report the average values across the entire S-subset.

\paragraph{P-subset (Performance)}  
The P-subset evaluates the performance of the OS agent. We report the following metrics: \textbf{Type} accuracy, which is the accuracy of the type of action the agent produces; 
Step-wise Success Rate (\textbf{SR}), which is the success rate at each individual step; 
and Trajectory Success Rate (\textbf{TSR}), which is the success rate across entire trajectories.
We report these performance metrics for different difficulty levels in the P-subset and for the entire P-subset.

\paragraph{E-subset (Efficiency)}  
The E-subset evaluates the efficiency of OS agents. 
We report the following metrics: the average number of input tokens per step, the average number of output tokens per step, and the average execution time per step. 
These efficiency metrics are compared with the SR achieved by the agent.

\paragraph{R-subset (Robustness)}  
The R-subset evaluates the robustness of OS agents. We report the SR of OS agents under various conditions: \textbf{Normal} (no perturbation applied), \textbf{Mask} (random masking of elements in the screenshot), \textbf{Zoom-In} (cropping of the screenshot), \textbf{30\_Gauss}, \textbf{50\_Gauss}, \textbf{70\_Gauss} (Gaussian noise applied to the screenshot at varying levels), \textbf{SC} (a signal indicating that the task has already been completed), \textbf{BM} (a signal indicating a bad memory), \textbf{BK} (a signal indicating bad knowledge), \textbf{IM} (a signal indicating an irrelevant memory), and \textbf{IK} (a signal indicating irrelevant knowledge).
These metrics comprehensively evaluate the robustness of OS agents.

\paragraph{Overall Ranking}
We report the rankings of each OS agent across different subsets and their overall ranking. 
The calculation rules are as follows: 
For the S-subset, we rank based on the Gold-Dist. score, with higher scores indicating better rankings; 
for the P-subset, we rank according to the average TSR, with higher scores indicating better rankings; 
for the E-subset, we rank based on the token and inference time separately for the OS agent, and then average the two rankings. The ranking for tokens is calculated as the sum of the average input tokens and three times the average output tokens, with a smaller value resulting in a higher rank (following the existing token billing methods used by model vendors such as OpenAI).
for the R-subset, we measure the relative decrease in performance compared to the Normal condition across ten types of perturbations, with smaller decreases indicating higher rankings.
The overall ranking is then computed by averaging the rankings across the four subsets.

\subsection{Main Results}

\begin{table*}[t]
\renewcommand{\arraystretch}{1.2}
\centering
\caption{Experimental results on the S-subset under different unsafe scenarios. Gold represents the proportion of gold actions that avoid unsafe factors, Dist. refers to the proportion of actions that do not avoid unsafe factors, and Inv. is the proportion that does not fall under the previous two categories.}
\label{table:S-subset}
\setlength{\tabcolsep}{5pt}
\begin{tabular}{ccccccccccccc}
\toprule
\multirow{2}{*}{\textbf{OS Agent}} & 
\multicolumn{3}{c}{\makecell{\bfseries Environmental \\ \bfseries Distractions}} & 
\multicolumn{3}{c}{\makecell{\bfseries Real-world \\ \bfseries Anomalies}} & 
\multicolumn{3}{c}{\makecell{\bfseries Adversarial \\ \bfseries Misleading}} & 
\multicolumn{3}{c}{\textbf{Avg.}} \\
\cline{2-13}
& Gold & Dist. & Inv. & Gold & Dist. & Inv. & Gold & Dist. & Inv. & Gold & Dist. & Inv. \\
\midrule
UI-TARS-2B-SFT & 78.11& 3.89 & 18.01& 51.06 & 48.94 & 0.00&  63.68& 6.60&  29.72& 67.91& 7.40& 24.68 \\
UI-Venus-1.5-2B & 59.46 & 3.24 & 37.31 & 34.04 & 65.96 & 0.00& 77.83 & 5.46& 16.71 & 70.04 & 7.15 & 22.81\\
Qwen3-VL-2B & 47.15 & \textbf{0.91} & 51.94 & 39.36 & 60.64 & 0.00 & 58.15 & 8.36& 33.49 & 53.79 & 8.00 & 38.21 \\
MAI-UI-2B & 30.96& 1.17 & 65.41& 15.96 & 77.66 & 0.00& 64.08 & 7.28 & 23.72 &51.28 & 8.09& 36.47\\
Qwen2.5-VL-3B & 80.05 & 4.27 &15.67 & 36.17 &63.83 & 0.00 & 55.19 &13.01 & 31.81 &62.60& 12.17& 25.23 \\
GELab-Zero-4B & 85.23 & 1.68 & 13.08 & 48.94 & 51.06 & 0.00 & 69.00 &4.04 & 26.95 & 73.53 & \textbf{5.15} & 21.32 \\
Qwen3-VL-4B & 68.01 & 2.98 & 29.02 & 48.94 &51.06 & 0.00 & 65.09 &12.60 & 22.30 & 65.40 & 10.98 & 23.62 \\
OS-Atlas-Pro-7B & 54.27& 6.99& 38.73& 45.74 &54.26 & 0.00 &62.87  & 7.68&29.45  &59.36 &9.32 & 31.32 \\
UI-TARS-7B-SFT &80.70 & 8.16 & 11.14 & 57.45 & 42.55 & 0.00&  71.02 & 6.06 & 22.91 & 73.66 & 8.21 &  18.13\\
Qwen2.5-VL-7B & 66.45 & 4.53 & 29.02 & 36.17 & 63.83 & 0.00 & 64.62 & 7.21 & 28.17 & 64.09 & 8.60 & 27.32 \\
GUI-owl-7B & 81.48 & 3.24 & 15.28 & 41.49 & 58.51 & 0.00 &  64.62 & 8.56 & 26.82 & 69.23 & 8.81 & 21.96  \\
UI-TARS-1.5-7B & 87.05 & 1.68 &11.27 & \textbf{63.83} & \textbf{36.17} & 0.00 &  66.24 & 7.41 & 26.35 &72.98 &6.68 &20.34   \\
AgentCPM-GUI-8B & 75.65 & 4.79 & 19.56 & 27.66 &72.34 & 0.00& 59.16 & 3.98 & 36.86 & 63.32 & 6.98 &  29.70 \\
UI-Venus-1.5-8B & 82.90 & 4.79 & 12.31 & 38.30 & 61.70 & 0.00 & \textbf{80.93} & \textbf{3.91} & 15.16 & 79.87 & 6.51& 13.62 \\
Qwen3-VL-8B & 81.99 & 3.63 & 14.38 & 45.74 &54.26  & 0.00 & 61.19 & 15.90 & 22.91 & 67.40 & 13.40 & 19.19  \\
MAI-UI-8B  & 79.15 & 1.42 & 17.75& 37.23 & 61.70& 0.00 & 66.64 & 7.35 & 25.40 & 69.57 & 7.57 & 21.87  \\
Qwen2.5-VL-32B  & 71.63 & 3.63 & 24.74& 25.53 & 74.47 & 0.00 & 76.42 & 6.87 & 16.71 & 72.81 & 8.51 &  18.68 \\
GUI-owl-32B & 84.46 & 2.85 & 12.69 & 27.66 & 72.34 & 0.00 & 68.80 & 6.67 & 24.53 & 72.30 & 8.04 & 19.66 \\       
Qwen3-VL-32B  & 81.87 & 3.24 & 14.90& 48.94 & 51.06 & 0.00 & 72.17 & 10.24 & 17.59 & 74.43 & 9.57&  16.00 \\
UI-TARS-72B-SFT & \textbf{89.12} & 4.27 & \textbf{6.61} & 44.68 & 55.32 & 0.00 & 67.86 & 5.46 & 26.68 & 73.91 & 7.06 &  19.02 \\
Qwen2.5-VL-72B & 76.94& 3.37 &19.69 & 37.23 & 62.77 & 0.00& 68.19 & 4.72 & 27.09 & 69.83 & 6.60 & 23.57   \\
GLM-4.5V &86.66 & 4.02&9.33 & 54.26& 45.74&0.00 & 79.25 &11.32 &\textbf{9.43} &\textbf{80.68} & 10.30& \textbf{9.02}  \\
\bottomrule
\end{tabular}
\end{table*}

\begin{figure}[t]
    \centering        
    \includegraphics[width=0.5\textwidth]{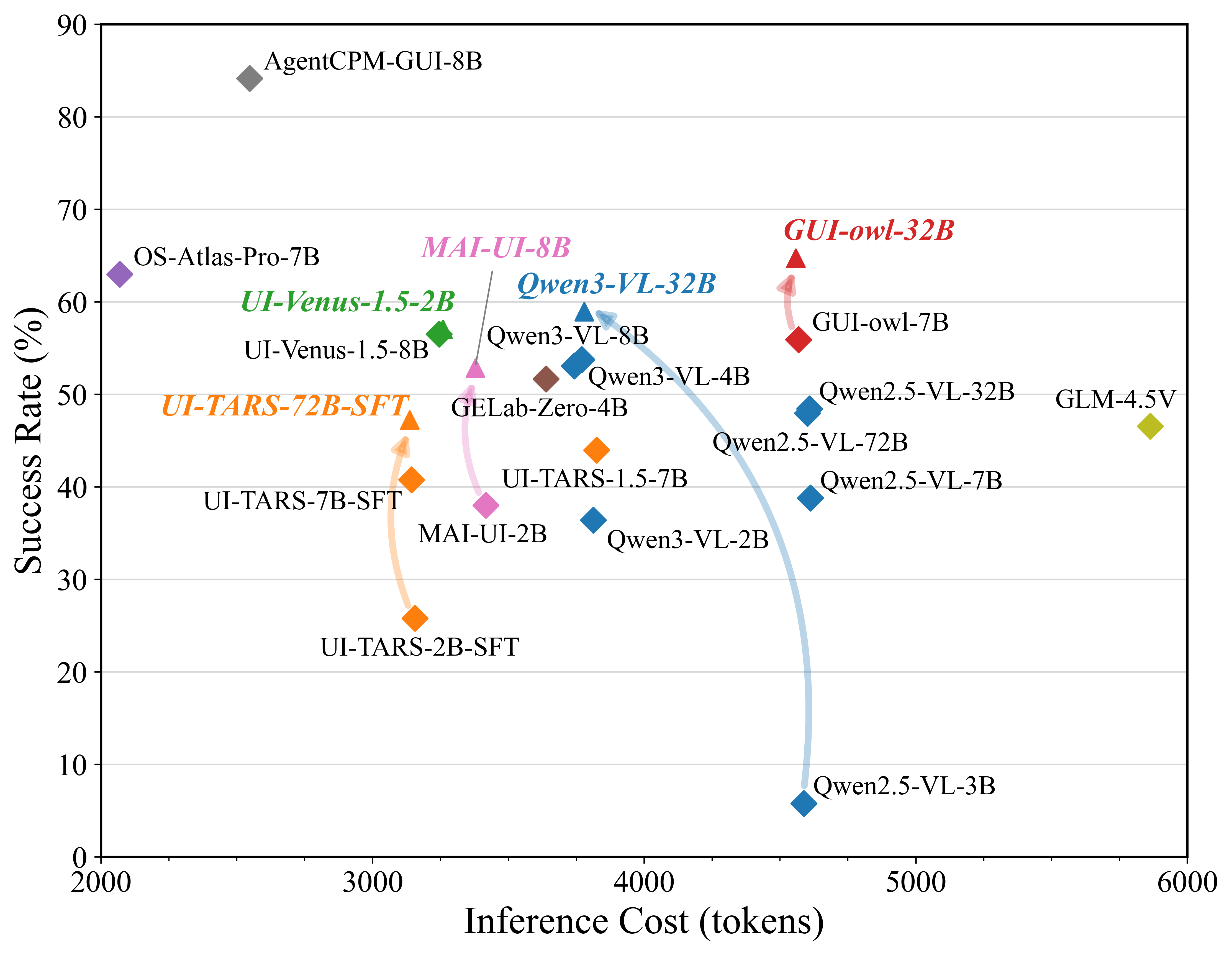} 
    \caption{Evaluation results of different OS agents in the E-subset's token dimension. The calculation of Inference cost (tokens) is based on the total input tokens plus three times the total output tokens. Models within the same series are connected by colored arrows.}
    \label{e-subset-token} 
\end{figure}

\begin{figure}[t]
    \centering        
    \includegraphics[width=0.5\textwidth]{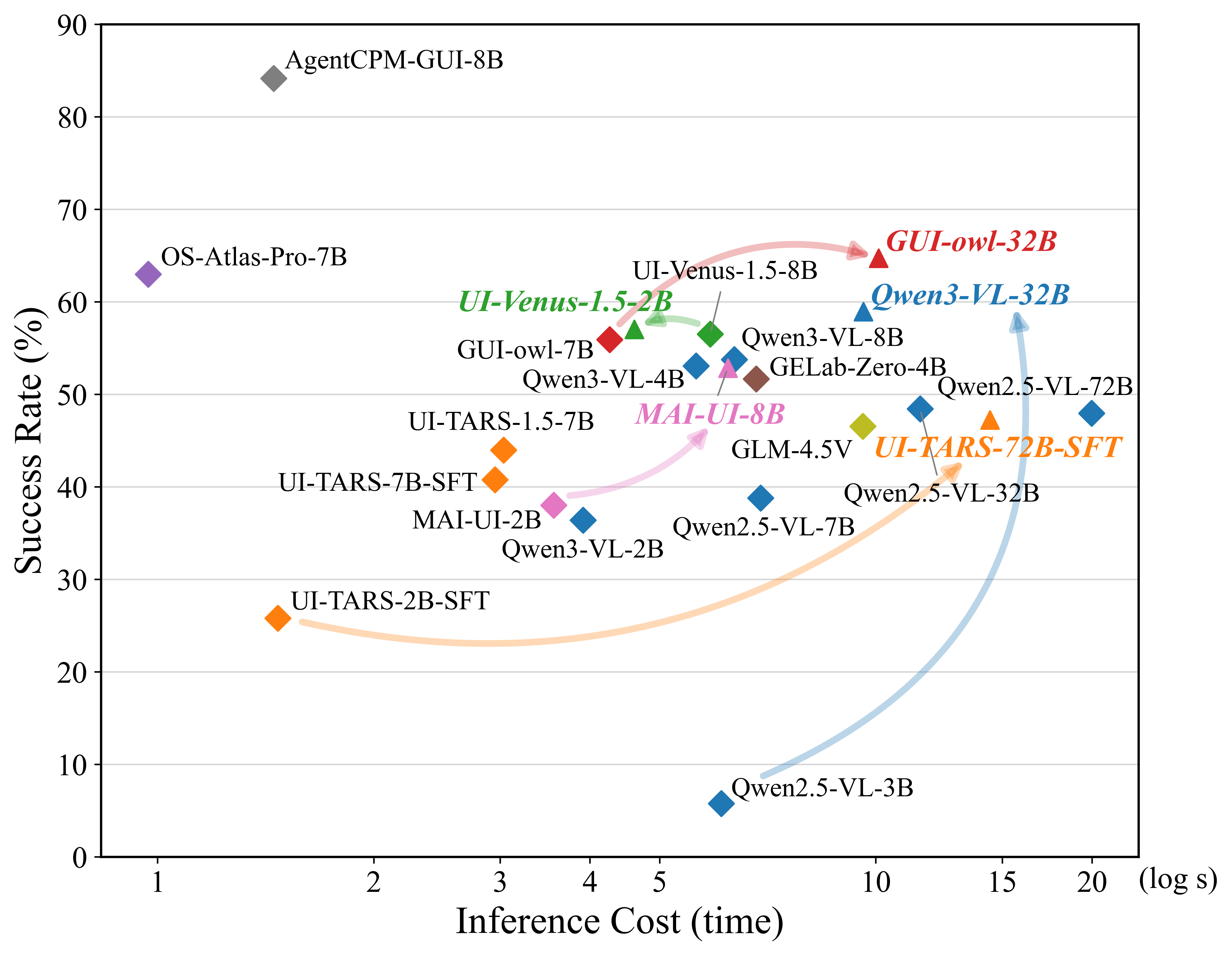} 
    \caption{Evaluation results of different OS agents in the E-subset's time dimension. Models within the same series are connected by colored arrows.Note that the horizontal axis is in logarithmic scale, with time measured in seconds.}
    \label{e-subset-time} 
\end{figure}
\begin{table*}[t]
\renewcommand{\arraystretch}{1.2}
\centering
\caption{Experimental results on P-subset under different difficulty levels. Type, SR and TSR are reported in (\%).}
\label{table:P-subset}
\setlength{\tabcolsep}{4.5pt}
\begin{tabular}{ccccccccccccc}
\toprule
\multirow{2}{*}{\textbf{OS Agent}} & \multicolumn{3}{c}{\textbf{Easy}} & \multicolumn{3}{c}{\textbf{Medium}} & \multicolumn{3}{c}{\textbf{Hard}} & \multicolumn{3}{c}{\textbf{Avg.}} \\
\cline{2-13}
& Type & SR & TSR & Type & SR & TSR & Type & SR & TSR & Type & SR & TSR \\
\midrule
UI-TARS-2B-SFT & 44.62 &28.86 &8.63 &35.41 &21.69 &3.50 &42.77 &26.19 &0.55 &41.11 &25.79 &5.03 \\ 
UI-Venus-1.5-2B & 78.94 & 64.26 & 11.82 & 71.55 & 50.74 & 8.50 & 73.20 & 53.82& 4.92 & 74.97 & 57.02 & 9.05 \\ 
Qwen3-VL-2B & 54.74 &40.33 & 6.07& 51.10 &32.32 &4.00 & 51.59&35.46 &1.09 & 52.70 & 36.40 & 4.17\\ 
MAI-UI-2B & 40.96 & 29.95 & 7.99& 34.30 & 19.66 &1.50 & 38.08 &23.79 &  0.55& 38.01& 24.92& 4.17 \\ 
Qwen2.5-VL-3B & 55.49 & 41.00 & 7.03 & 47.84 & 29.05 & 6.50 & 52.60 & 36.13 & 1.64 & 52.21 & 35.78& 5.46 \\
GELab-Zero-4B & 73.15 & 57.58 & 13.10 & 66.16 & 45.35 & 5.50 & 69.63 & 50.47& 7.10 & 69.91 & 51.66 & 9.34 \\
Qwen3-VL-4B & 76.89 & 60.53 & 15.65& 66.02 &44.48 & 9.50& 71.08&52.32 &8.20 & 71.77& 53.06& 11.93\\ 
OS-Atlas-Pro-7B & 77.93 & 65.98 & 13.10 & 78.38 & 61.65 & 10.89 &77.33  &60.08 &8.74 & 77.92 & 62.97 & 11.32  \\
UI-TARS-7B-SFT & 60.46 & 45.67 & 10.54 & 51.15 & 35.45 & 8.50& 55.61& 39.92& 5.46& 56.11& 40.78&8.62 \\
Qwen2.5-VL-7B & 58.44 & 45.63 & 11.50 & 49.86 & 31.22 & 6.50 & 55.39 & 37.74 & 2.19 & 54.81 & 38.79& 7.61\\
GUI-owl-7B & 78.57 & 61.73 & 15.02 & 71.55 & 51.34 &13.00 & 71.73 & 52.76 & 7.65  & 74.27 & 55.91 & 12.50  \\
UI-TARS-1.5-7B & 72.63&50.41 &11.18 &63.17 & 36.79& 9.00& 66.39 & 43.10 & 3.28 & 67.85 & 43.98 &8.48 \\
AgentCPM-GUI-8B & \textbf{94.70} & \textbf{88.16} & \textbf{38.02}& \textbf{93.05} & \textbf{82.46} &\textbf{12.50} &\textbf{90.68} & \textbf{80.18} & \textbf{19.13} & \textbf{93.07} & \textbf{84.14}& \textbf{25.72}\\
UI-Venus-1.5-8B &77.11 &64.19  & 14.38 & 66.62 & 48.30 & 14.50 & 71.69 & 55.00 & 7.65 & 72.22 & 56.51 & 12.64  \\
Qwen3-VL-8B & 78.27 & 61.28 &17.57 &67.31 &45.35 &13.00 &70.24 & 52.71& 3.83& 72.52 & 53.76& 12.64\\ 
MAI-UI-8B &56.16 &44.77 & 10.54& 48.30& 31.58& 5.50& 53.38& 37.41& 3.28& 52.84& 38.47&7.18 \\  
Qwen2.5-VL-32B & 65.20 & 54.14 & 10.54 & 59.12 & 44.15 & 7.50 & 61.59 & 45.06& 5.46 & 62.23 & 48.43 & 8.33 \\
GUI-owl-32B & 83.08 & 70.43 & 17.89 & 79.60 & 65.61 & 21.00 & 73.26 & 55.11 & 18.03 & 79.30 & 64.72 & 18.82 \\
Qwen3-VL-32B & 81.81 & 67.21 & 22.04 & 68.83 & 48.62 & 13.50 & 75.49 & 59.07 & 9.29 & 75.86 & 58.94 & 16.24 \\ 
UI-TARS-72B-SFT & 65.05 & 53.10 & 15.02 & 52.07 & 40.93 & 15.00 & 59.24 & 46.18 & 8.74 & 59.24 & 47.25 & 13.36\\ 
Qwen2.5-VL-72B & 65.46 & 52.39 & 9.58& 60.27 &44.11 &6.50 &63.60 &46.01 &6.56 &63.26 & 47.96 & 7.90\\
GLM-4.5V & 72.89&52.84 &12.14 & 63.58 &38.03 & 9.50 & 68.57&47.46 &6.01 &68.68 &46.54 &9.77\\
\bottomrule
\end{tabular}
\end{table*}

\begin{table*}[t]
\renewcommand{\arraystretch}{1.2}
\centering
\caption{Experimental results on R-subset under different robustness perturbations. SR is reported in (\%). \\Abbreviations: SC = State Conflict, BM = Bad Memory, BK = Bad Knowledge, IM = Irrelevant Memories, IK = Irrelevant Knowledge.}
\label{table:R-subset}
\setlength{\tabcolsep}{3.5pt}
\begin{tabular}{cccccccccccc}
\toprule
\textbf{OS Agent} & \textbf{Normal} & \textbf{Mask} & \textbf{Zoom\_In} & \textbf{30\_Gauss} & \textbf{50\_Gauss} & \textbf{70\_Gauss} & \textbf{SC} & \textbf{BM} & \textbf{BK} & \textbf{IM} & \textbf{IK} \\
\midrule
UI-TARS-2B-SFT & 45.80& 42.20& 44.00& 46.00& 45.30& 43.70& 49.70& 41.90& 47.90& 46.90&25.10\\
UI-Venus-1.5-2B & 64.70 & 58.10& 60.10&61.40 &60.40 &59.60 &65.40 &66.30 & 67.40&64.40 &47.50\\
Qwen3-VL-2B & 42.00 & 41.10 & 44.30& 39.90 &40.90 &40.20 &36.40 &43.70 & 56.20& 38.50&31.60\\
MAI-UI-2B & 38.10& 27.20& 30.20& 35.40& 34.30& 34.10& 36.60& 47.00& 39.90&40.00 &21.80\\
Qwen2.5-VL-3B  & 60.00& 46.20 &53.30 &58.00 &55.60 &54.90 &55.10 &61.10 &55.20 &57.20 &46.10\\
GELab-Zero-4B &62.80 &55.10&55.40 &61.70 & 60.10& 59.00& 58.60& 63.80& 69.80&62.80 &60.30 \\
Qwen3-VL-4B & 63.00 & 59.40 & 59.00 & 60.50 & 60.80 & 59.80& 55.00& 66.30& 71.40& 61.50&56.10\\
OS-Atlas-Pro-7B & 56.00& 49.00&50.60 &54.60 & 52.80& 53.60 & 54.80& 62.50 & 56.90 &47.20&36.90\\
UI-TARS-7B-SFT & 55.40& 47.00 & 52.80 & 54.80 & 55.20 & 52.00 & 55.60 & 57.60& 59.40&55.60 &51.10\\
Qwen2.5-VL-7B &61.10 & 58.40&57.40 &60.50 &57.80 &56.90 &60.90 &64.70 &65.70 &61.50 &55.10\\
GUI-owl-7B & 62.10 & 56.60 &62.70 &60.30 &60.30 &59.10 &57.80 &65.00&68.60 &61.80 &57.70\\
UI-TARS-1.5-7B & 56.60 &47.80 &50.20 &56.10 &53.90 & 53.40& 56.30&56.70 &60.20 &54.10 &40.80\\
AgentCPM-GUI-8B & 62.50 & 50.50 & 51.40& 63.00& 62.20 & 62.60 & 59.00 & 61.50 & 70.40& 61.50&58.10\\
UI-Venus-1.5-8B & \textbf{68.40} & 61.60& 59.60& \textbf{67.50}& \textbf{66.90} & \textbf{65.30}& \textbf{69.20} & \textbf{73.40}& 72.60&\textbf{68.60} &56.90\\
Qwen3-VL-8B & 62.80& 59.10&57.50 &60.60 &59.50 &58.50 &34.90 &65.80 &70.60 &62.10 &35.30\\
MAI-UI-8B &52.80 & 46.40& 47.10& 50.00& 49.60 & 49.70 &57.30 & 61.70 & 56.20& 54.10&46.70\\
Qwen2.5-VL-32B & 65.80 & 60.30 & \textbf{65.40}&64.60 &64.30 & 62.70& 57.10& 67.80&70.60 &66.00 &\textbf{61.60}\\
GUI-owl-32B  &62.40 & 56.20 & 62.20 & 63.80 & 61.40 & 62.60 & 60.40 & 71.40&73.00 &64.70 &53.80\\
Qwen3-VL-32B & 67.00 & \textbf{63.20} & 62.70& 65.60 &64.50 &63.60 &46.90 & 67.50&\textbf{74.10}&64.50&50.10\\
UI-TARS-72B-SFT & 56.90 & 44.90 & 52.30& 55.70& 54.40& 53.60& 60.90& 65.10&63.30 &57.70 &46.00\\
Qwen2.5-VL-72B & 67.40 & 56.60& \textbf{65.40}& 63.30& 62.40& 59.40& 65.10& 71.10& 68.90& 65.70&61.30\\
GLM-4.5V & 58.00 & 57.20 & 56.60 &54.40 &58.00 &58.00 &46.80 & 58.40 &64.70 &57.80 &48.90 \\
\bottomrule
\end{tabular}
\end{table*}

The overall ranking of different OS agents in OS-SPEAR is shown in Table~\ref{table:total}, while their detailed performance across different subsets is provided in Tables~\ref{table:S-subset}, \ref{table:P-subset}, \ref{table:R-subset}, Figure~\ref{e-subset-token} and Figure~\ref{e-subset-time}.
We comprehensively compared the capability differences among different OS agents using these data.

Overall, UI-Venus-1.5-8B achieved the first place in OS-SPEAR. This is because UI-Venus-1.5-8B demonstrated excellent task completion capability (5th in the P-subset), along with outstanding safety performance (1st in the S-subset), decent efficiency (7th in the E-subset), and robustness (10th in the R-subset).
Other OS agents, however, tend to have certain weaknesses in different dimensions. For example, GUI-owl-32B performs excellently in safety, performance, and robustness, but due to its large parameter size (32B) and the high number of generated thoughts, it struggles in terms of efficiency, which negatively affects its overall ranking.
Similarly, AgentCPM-GUI-8B balances performance (1st in the P-subset) and efficiency (2nd in the E-subset), but its performance in task completion for OS operations is somewhat compromised by hacking issues and insufficient reasoning chains, resulting in poor safety (15th in the S-subset) and robustness (13th in the R-subset).

The experimental results from OS-SPEAR have led to many exciting findings, which we will discuss in Section~\ref{key_findings}.

\section{Key Findings}
\label{key_findings}
Based on OS-SPEAR, we have obtained many exciting findings. 
In this section, we will sequentially introduce the key findings in the S-subset, P-subset, E-subset, and R-subset, followed by the overall insights from OS-SPEAR.
\subsection{Key Findings of S-subset}

\paragraph{\textbf{General-purpose models with OS capabilities exhibit weaker safety than specialized OS agents}}
As shown in Table~\ref{table:S-subset} and Table~\ref{table:total}, only one general-purpose model (GLM-4.5V) ranks within the top six on the S-subset, while five of the bottom seven models are general-purpose models. 
This is primarily because general-purpose models possess stronger semantic understanding, making them more susceptible to recognizing and following malicious or injected intentions in the environment, which can lead to unsafe actions.

\paragraph{\textbf{Real-world anomalies are harder to mitigate than adversarial attacks}}
As shown in Table~\ref{table:S-subset}, OS agents achieve significantly lower gold action rates when facing real-world anomalies compared to environmental distractions or adversarial misleading inputs. 
This is because artificially injected unsafe factors often deviate from the natural semantic distribution of the environment, whereas real-world anomalies are inherently embedded within it, making them more difficult to distinguish and thus more misleading.

\paragraph{\textbf{Larger models tend to exhibit better safety performance}}
As shown in Table~\ref{table:S-subset} and Table~\ref{table:total}, all six OS agents with parameter sizes greater than or equal to 32B rank within the top ten on the S-subset. 
This suggests that larger models are better at distinguishing between useful and unsafe information, enabling more reliable decision-making.

\subsection{Key Findings of P-subset}

\paragraph{\textbf{Higher SR does not necessarily imply higher TSR}}
As shown in Table~\ref{table:P-subset}, models with higher Step Success Rate (SR) do not always achieve higher Task Success Rate (TSR). 
For example, OS-Atlas-Pro-7B achieves approximately 10\% higher SR than Qwen3-VL-4B but yields a lower TSR. 
Trajectory analysis reveals that such models tend to fail at specific critical steps (e.g., prematurely outputting a termination action), leading to early task termination despite strong step-level performance.

\paragraph{\textbf{Scaling improves OS agent performance}}
As shown in Table~\ref{table:P-subset} and Table~\ref{table:total}, within model families such as UI-TARS, Qwen2.5-VL, and Qwen3-VL, both SR and TSR consistently improve with increasing parameter size. 
This indicates that scaling laws are also applicable to OS agents.

\subsection{Key Findings of E-subset}

\paragraph{\textbf{Higher inference cost does not guarantee better performance}}
As shown in Figure~\ref{e-subset-time} and Figure~\ref{e-subset-token}, there is no clear linear correlation between inference cost and SR across different OS agents. 
This suggests that simply increasing context length or test-time computation does not necessarily lead to improved task performance.

\paragraph{\textbf{Within the same model family, gains from increased inference cost are limited}}
As shown in Figure~\ref{e-subset-token}, models within the same family have similar token-level inference costs due to shared input formatting and output structures. 
Although scaling increases time-level inference cost (Figure~\ref{e-subset-time}), the corresponding SR improvement is relatively modest compared to switching to a different model family.

\subsection{Key Findings of R-subset}

\paragraph{\textbf{OS agents heavily rely on the completeness of visual inputs}}
As shown in Table~\ref{table:R-subset}, perturbations such as masking and zooming—both of which disrupt screenshot completeness—lead to significant performance degradation, even though the ground-truth regions remain intact. 
In contrast, Gaussian noise, which preserves structural completeness, has a much smaller impact on performance.

\paragraph{\textbf{OS agents are insensitive to temporal causality in textual inputs}}
As shown in Table~\ref{table:R-subset}, when shuffled bad memory or injected knowledge is provided, OS agents can still complete tasks effectively, sometimes even with improved performance. 
This indicates that OS agents can extract key information without relying on temporal order in textual inputs.

\paragraph{\textbf{OS agents are vulnerable to conflicting and irrelevant textual information}}
As shown in Table~\ref{table:R-subset}, perturbations such as state conflicts, irrelevant memories, and irrelevant knowledge significantly degrade performance across most models. 
This suggests that OS agents are easily influenced by certain textual signals, which can mislead action generation.

\paragraph{\textbf{Robustness weaknesses vary across modalities}}
As shown in Table~\ref{table:R-subset} and Table~\ref{table:total}, robustness is modality-dependent. 
For instance, UI-TARS-72B-SFT performs well under textual perturbations but is vulnerable to visual disturbances, whereas GLM-4.5V shows the opposite trend.

\subsection{Overall Insights from OS-SPEAR}

\paragraph{\textbf{Efficiency often comes at the cost of safety and robustness}}
As shown in Table~\ref{table:S-subset}, Table~\ref{table:R-subset}, Figure~\ref{e-subset-time}, Figure~\ref{e-subset-token}, and Table~\ref{table:total}, models with lower inference cost (e.g., AgentCPM-GUI-8B and OS-Atlas-Pro-7B) tend to perform worse in safety and robustness. 
This is because efficiency-oriented designs often omit critical contextual reasoning, making these models more fragile under challenging conditions.

\paragraph{\textbf{Specialized OS agents outperform general-purpose models in overall performance}}
As shown in Table~\ref{table:total}, the top eight models in the overall ranking are all specialized OS agents. 
This demonstrates that domain-specific adaptation effectively removes redundant general-purpose capabilities, enabling more efficient, safe, and robust task execution in OS environments.
This shows that domain adaptation of general models for OS agents is meaningful.
Because during domain adaptation, OS agents not only improve their grounding capabilities but also enhance their understanding of OS operation logic, and acquire static knowledge about the OS (e.g., web or specific applications).

\paragraph{\textbf{Larger models within the same series do not always rank higher overall}} As shown in Table~\ref{table:total}, Qwen3-VL-4B ranks higher than Qwen3-VL-32B, and Qwen2.5VL-7B ranks higher than Qwen2.5VL-72B. This is because the performance scaling of models within the same series is limited. 
Simply scaling up the parameters of the OS agent without innovation in architecture, data, or algorithms will quickly hit a performance plateau. 
Additionally, scaling up can impose a greater burden on the efficiency dimension of the OS agent, leading to a decrease in the overall ranking.

\section{Conclusion}

As the paradigm of MLLMs shifts from passive text generation to active behavioral execution, the need for rigorous, multi-dimensional evaluation has never been more urgent. 
While existing benchmarks primarily focus on task completion rate.
Therefore, we present OS-SPEAR, a comprehensive toolkit designed to evaluate OS agents across the four critical dimensions of Safety, Performance, Efficiency, and Robustness. 
OS-SPEAR addresses the critical gap in assessing the reliability and trustworthiness of agents in complex, real-world environments.
We provide a holistic framework that broadens the current boundaries of autonomous OS agent capability evaluation.
By conducting an extensive benchmark of 22 OS agents, we revealed significant trade-offs between execution efficiency and safety/robustness, while identifying modality-specific vulnerabilities. 
Through its specialized subsets and automated diagnostic reporting, OS-SPEAR provides a rigorous and standardized framework that bridges the gap between task execution and real-world . 
We believe that by providing a standardized, multidimensional evaluation protocol, OS-SPEAR will serve as a foundational resource for the development of next-generation OS agents that are not only highly capable but also inherently safe, efficient, and resilient.
% \section*{Acknowledgment}

\ifCLASSOPTIONcaptionsoff
  \newpage
\fi

\bibliographystyle{IEEEtran}
\bibliography{manuscript}

\begin{IEEEbiography}[{\includegraphics[width=1in,height=1.25in,clip,keepaspectratio]{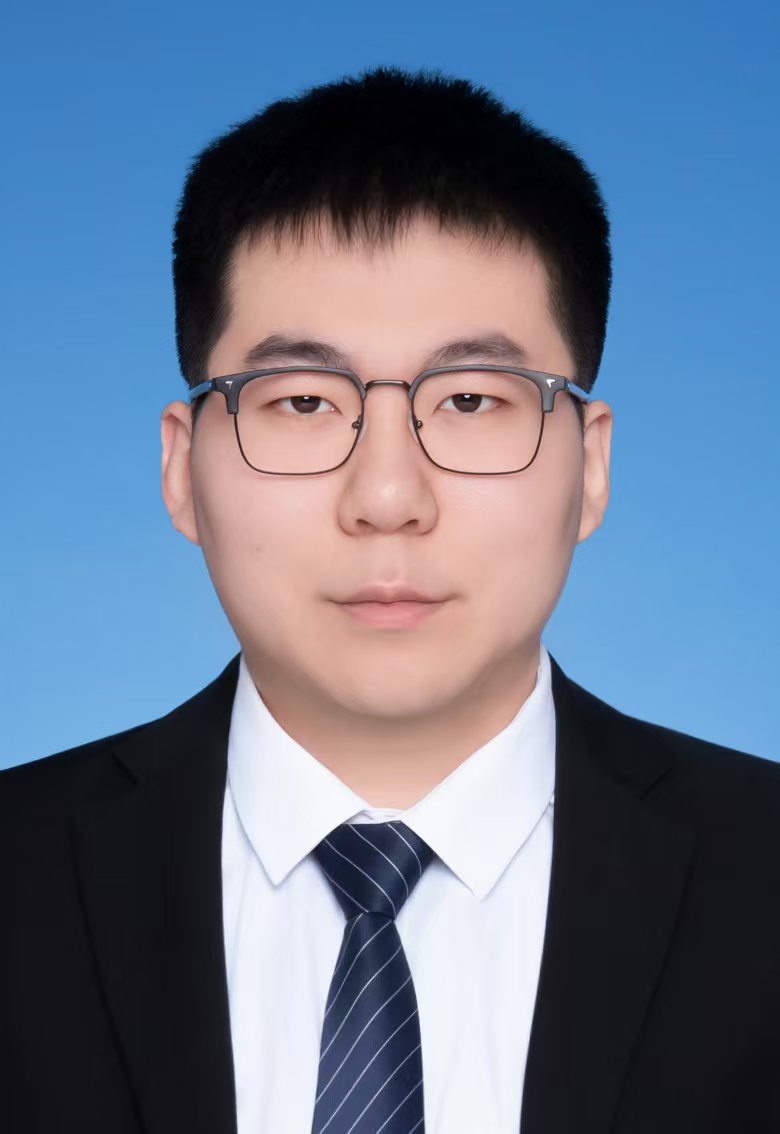}}]{Zheng Wu}
received his Bachelor's degree in information security from Shanghai Jiao Tong University, Shanghai, China, in 2025. He is working towards his M.S. degree at the AI Security Lab of Shanghai Jiao Tong University. His research interests include natural language processing, multimodal large language model, and AI agent.
\end{IEEEbiography}

\begin{IEEEbiography}[{\includegraphics[width=1in,height=1.25in,clip,keepaspectratio]{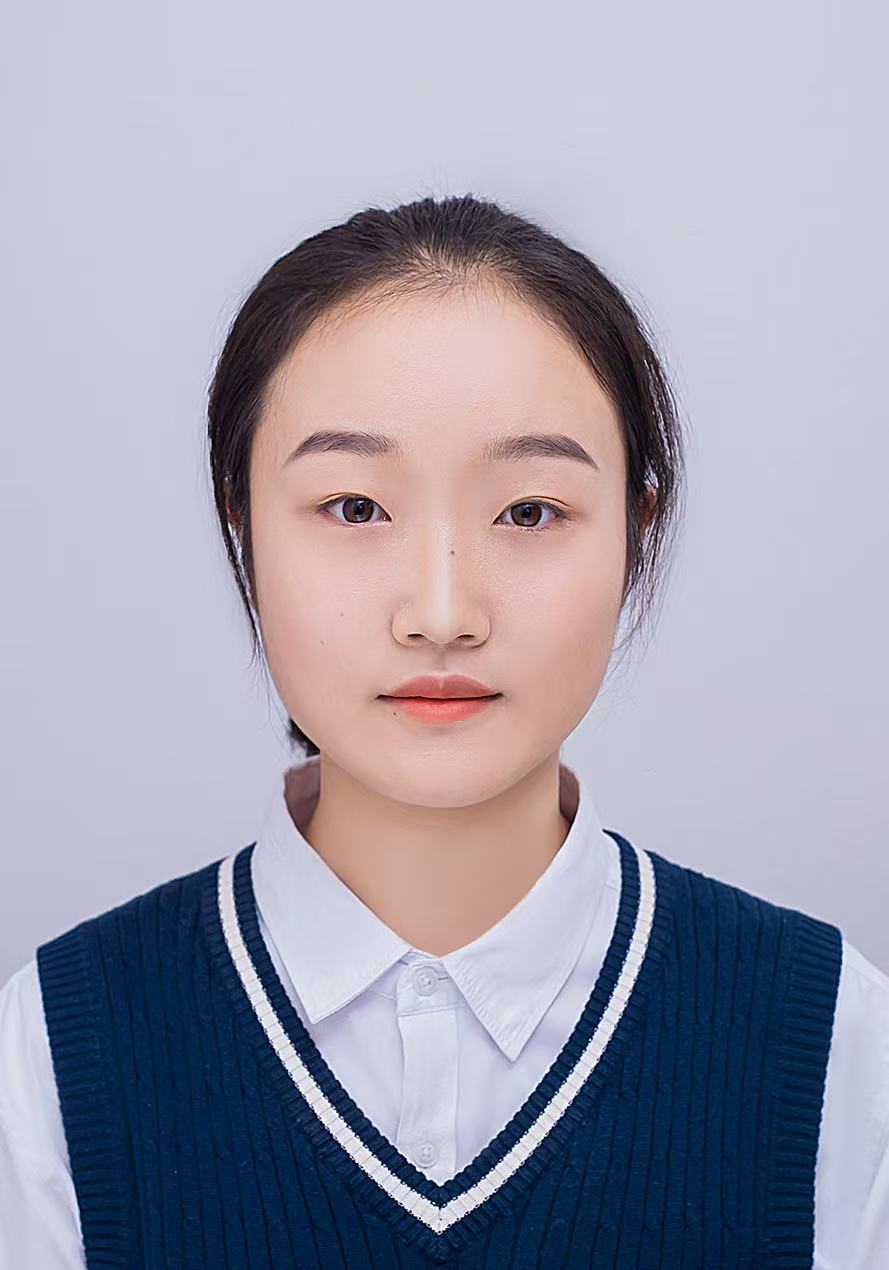}}]{Yi Hua}
will receive her Bachelor's degree in Information Security from Shanghai Jiao Tong University, Shanghai, China, in 2026. She is an incoming M.S. student at the AI Security Lab of Shanghai Jiao Tong University. Her research interests include natural language processing, large language models, and AI agents.
\end{IEEEbiography}

\begin{IEEEbiography}[{\includegraphics[width=1in,height=1.25in,clip,keepaspectratio]{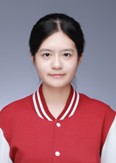}}]{Zhaoyuan Huang}
is currently an undergraduate student majoring in Information Security at Shanghai Jiao Tong University, Shanghai, China, and is expected to receive her Bachelor’s degree in 2027. She is a research intern at the AI Security Lab of Shanghai Jiao Tong University. Her research interests include natural language processing and the safety and robustness of AI agents.
\end{IEEEbiography}

\begin{IEEEbiography}[{\includegraphics[width=1in,height=1.25in,clip,keepaspectratio]{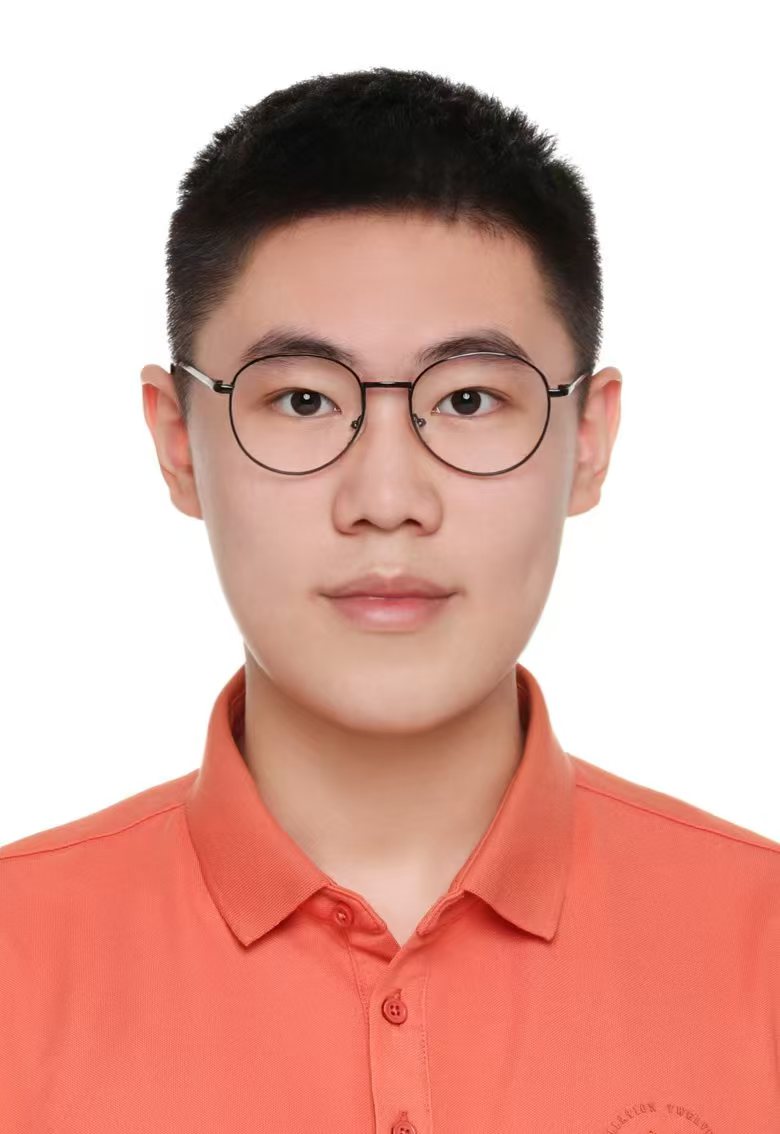}}]{Chenhao Xue} is currently an undergraduate student in the IEEE Pilot Class, majoring in Computer Science and Technology at Shanghai Jiao Tong University, Shanghai, China, and is expected to receive his Bachelor's degree in 2028. He is a research intern at the AI Security Lab of Shanghai Jiao Tong University. His research interests include data-centric AI, robustness of multimodal large language models, and autonomous agents.
\end{IEEEbiography}

\begin{IEEEbiography}[{\includegraphics[width=1in,height=1.25in,clip,keepaspectratio]{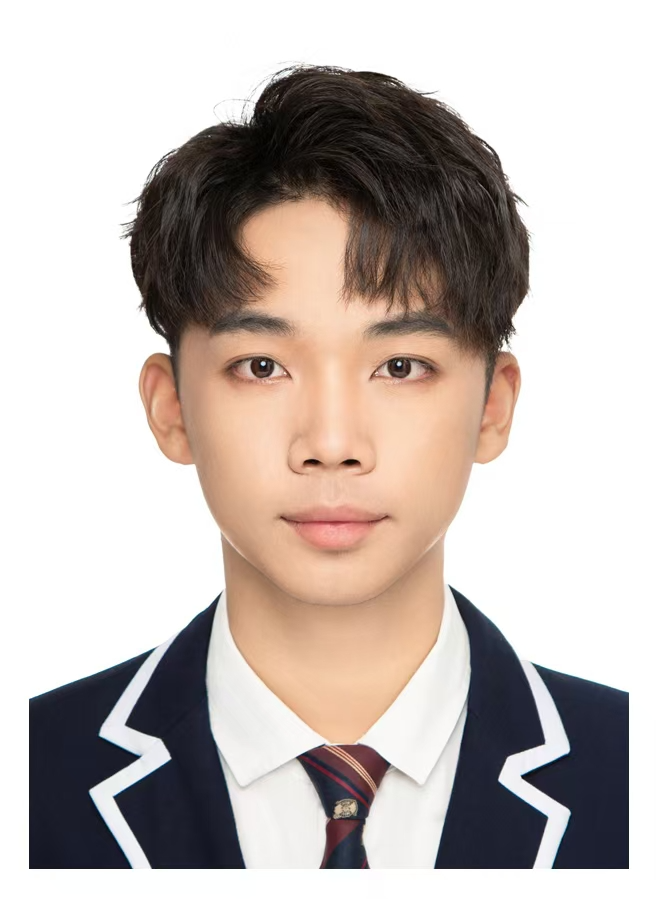}}]{Yijie Lu}
will receive his Bachelor's degree from Wuhan University in 2026. He is an incoming Ph.D. student at the AI Security Lab of Shanghai Jiao Tong University. His research interests include natural language processing, multimodal large language models, and agent security.
\end{IEEEbiography}

\begin{IEEEbiography}[{\includegraphics[width=1in,height=1.3in,clip,keepaspectratio]{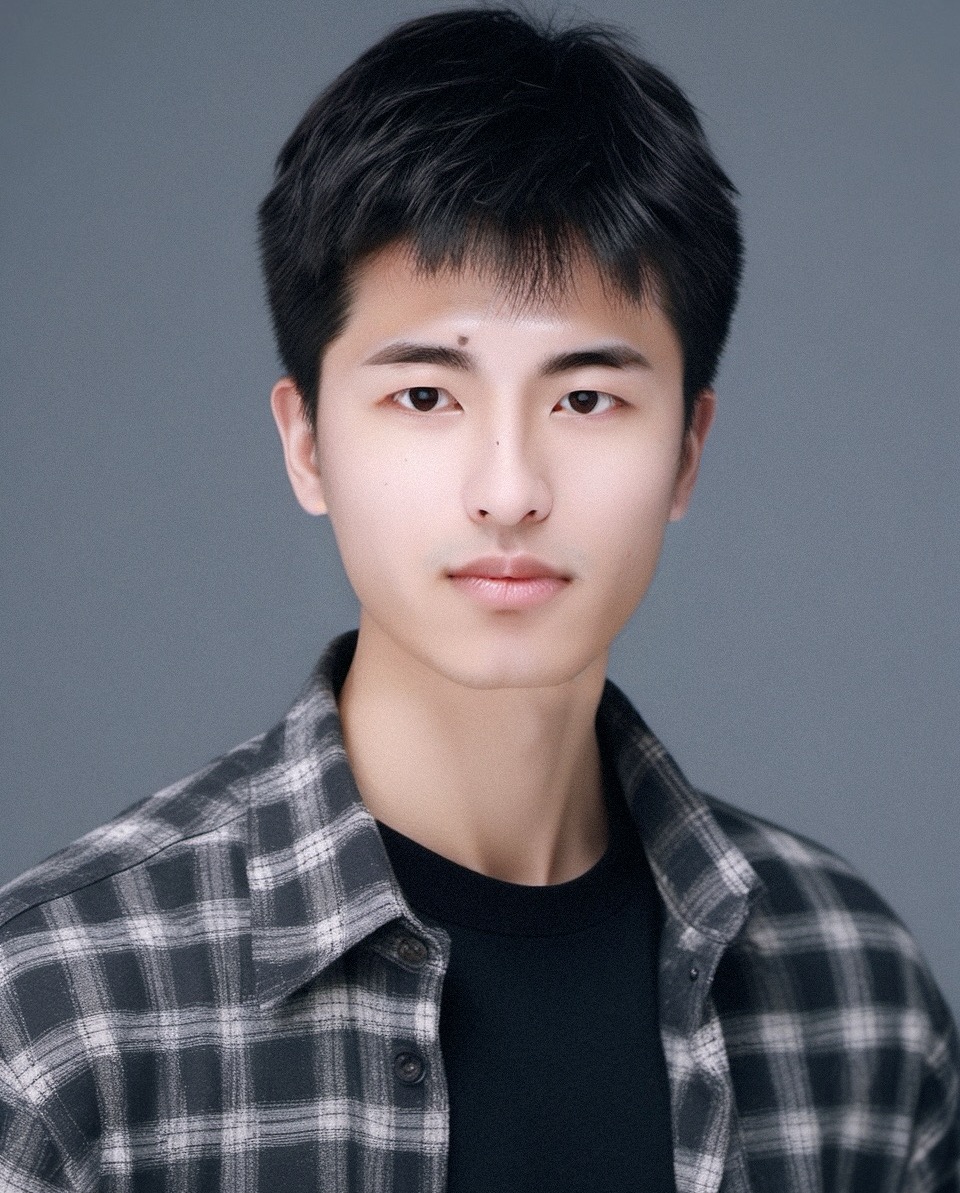}}]{Pengzhou Cheng}
received the M.S. Degree from the Department of Computer Science and Communication Engineering, Jiangsu University, Zhenjiang, China, in 2022. He is also co-supervised by Prof Fengwei Zhang of the Department of Computer Science and Engineering, Southern University of Science and Technology. He is currently pursuing the Ph.D. Degree with the Department of Electronic Information and Electrical Engineering, Shanghai Jiao Tong University, Shanghai, 201100, China. His primary research interests include LLM reasoning, AI Agent, artificial intelligence security, and cybersecurity.\end{IEEEbiography}

\begin{IEEEbiography}[{\includegraphics[width=1in,height=1.25in,clip,keepaspectratio]{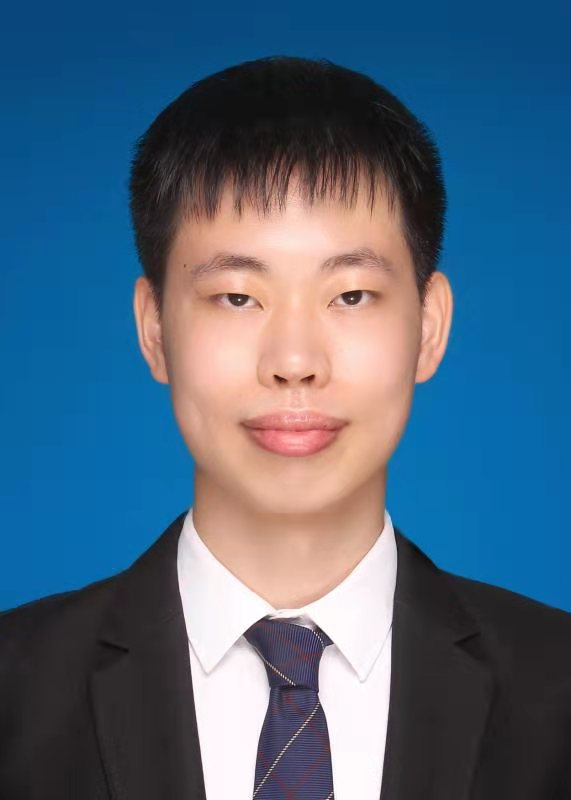}}]{Zongru Wu}
received the B.S Degree from the School of Cyber Science and Engineering, Wuhan University, Hubei, China, in 2022. He is currently persuing the Ph.D. Degree with the School of Cyber Science and Engineering, Shanghai Jiao Tong University, Shanghai, 201100, China. His primary research interests include LLM-powered agents, artificial intelligence security, backdoor attack and countermeasures, cybersecurity, and deep learning.
\end{IEEEbiography}

\begin{IEEEbiography}[{\includegraphics[width=1in,height=1.25in,clip,keepaspectratio]{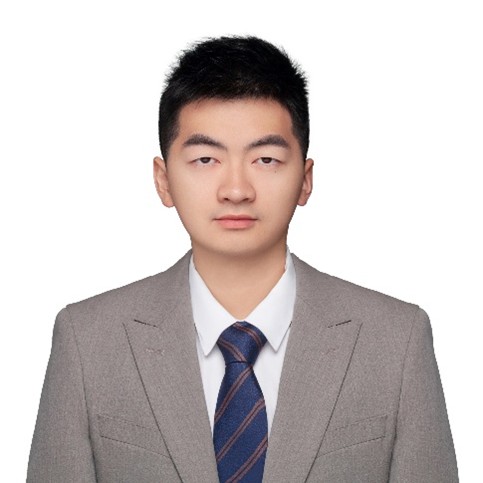}}]{Lingzhong Dong}
 received the B.S Degree from the School of Cyber Science and Engineering, Shanghai Jiao Tong University, Shanghai, China, in 2024. He is currently pursuing a Master's Degree with the School of Cyber Science and Engineering, Shanghai Jiao Tong University, Shanghai, 201100, China. His primary research interests include vision language models and graphical user interface agents.
\end{IEEEbiography}

\begin{IEEEbiography}
 [{\includegraphics[width=1in, height=1.25in, clip, keepaspectratio]{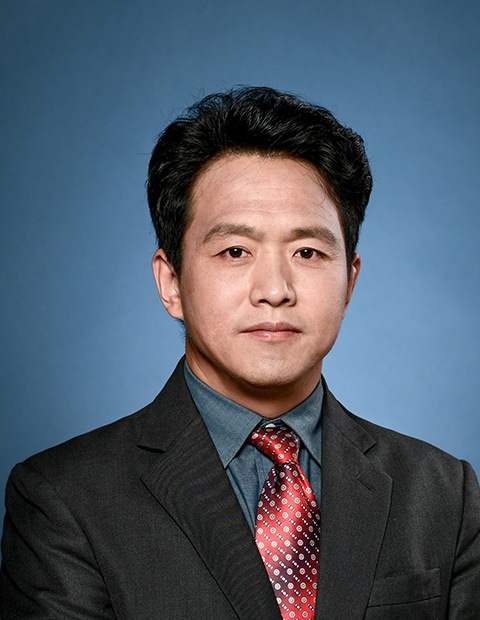}}]{Gongshen Liu}
received his Ph.D. degree in the Department of Computer Science from Shanghai Jiao Tong University in 2003. He is currently a professor with the School of Electronic Information and Electrical Engineering, Shanghai Jiao Tong University. His research interests cover natural language processing, machine learning, and artificial intelligence security.
\end{IEEEbiography}

\begin{IEEEbiography}[{\includegraphics[width=1in,height=1.25in,clip,keepaspectratio]{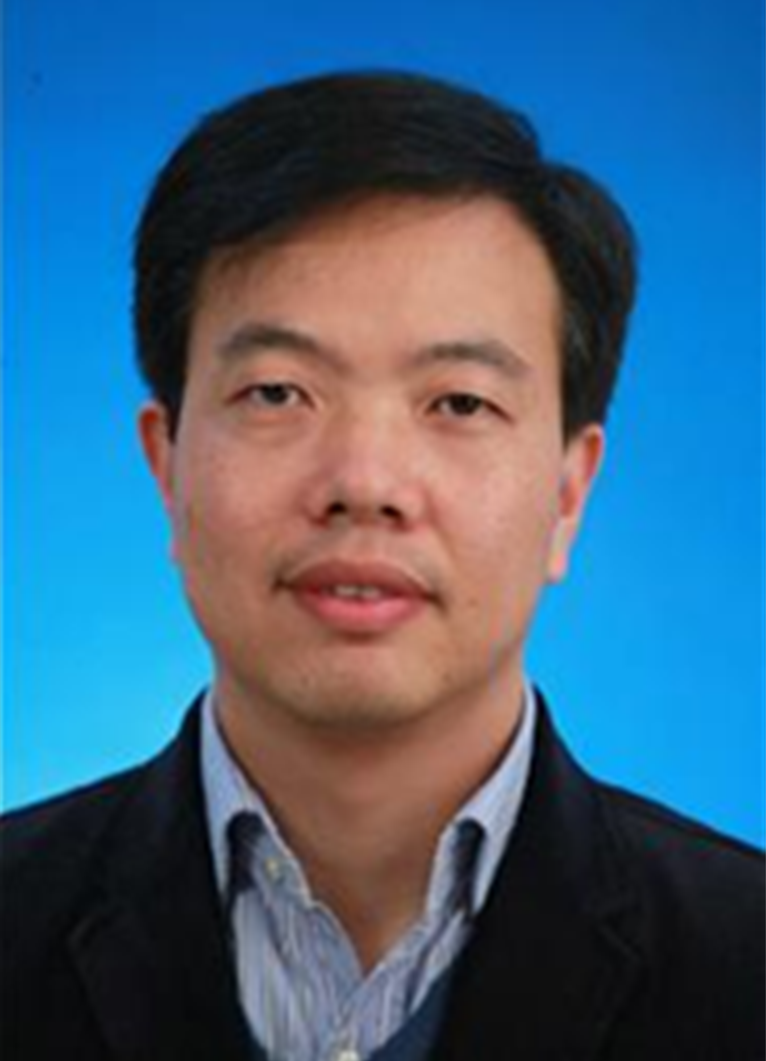}}]{Xinghao Jiang}
received the Ph.D. degree in electronic science and technology from Zhejiang University, Hangzhou, China, in 2003. He was a Visiting Scholar with the New Jersey Institute of Technology, Newark, NJ, USA, from 2011 to 2012. He is currently a Professor with the School of Computer Science, Shanghai Jiao Tong University, Shanghai, China. His research interests include multimedia security and image retrieval, intelligent information processing, cyber information security, information hiding and watermarking. Dr. Jiang is an IEEE senior member.
\end{IEEEbiography}

\begin{IEEEbiography}[{\includegraphics[width=1in,height=1.25in,clip,keepaspectratio]{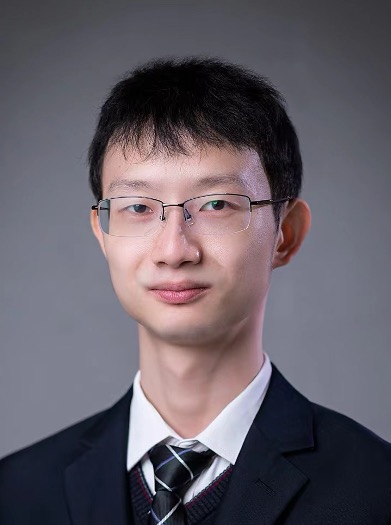}}]{Zhuosheng Zhang} received his Bachelor's degree from Wuhan University in 2016, and his M.S. and Ph.D. degrees from Shanghai Jiao Tong University in 2020 and 2023, respectively. He is currently a tenure-track assistant professor at Shanghai Jiao Tong University. He was a research intern at Amazon AWS, Microsoft Research, Langboat Technology, NICT (Japan), and IBM. His research interests include natural language processing, large language models, and language agents. 
He has published research papers in leading journals and conferences, such as TPAMI, TNNLS, TASLP, ICLR, ICML, ACL, AAAI, EMNLP, and COLING. 
He was the recipient of the WAIC 2024 Youth Outstanding Paper Award, WAIC 2024 YunFan Award, and the Global Top 100 Chinese Rising Stars in Artificial Intelligence. 
He serves as an action editor for ACL Rolling Review and standing reviewer for TACL. He served as a (senior) area chair for conferences such as NeurIPS, ACL, IJCAI, EMNLP, and COLING.
	\end{IEEEbiography}

\end{document}